\documentclass[10pt,twocolumn,letterpaper]{article}

\usepackage{cvpr}              
\usepackage{tabularx}
\usepackage{multirow}
\usepackage{xcolor}
\usepackage{algorithm}
\usepackage{algpseudocode}
\definecolor{cvprblue}{rgb}{0.21,0.49,0.74}
\usepackage[pagebackref,breaklinks,colorlinks,allcolors=cvprblue]{hyperref}

\def\paperID{4415} 
\def\confName{CVPR}
\def\confYear{2025}

\title{Timestep-Aware Diffusion Model for Extreme Image Rescaling}

\author{Ce Wang, Zhenyu Hu, Wanjie Sun\thanks{Corresponding author}, Zhenzhong Chen\\
School of Remote Sensing and Information Engineering, Wuhan University\\ Wuhan 430079, China\\
\tt\small {\{cewang, zhenyuhu, sunwanjie, zzchen\}@whu.edu.cn}
}

\begin{document}
\maketitle
\begin{abstract}
Image rescaling aims to learn the optimal low-resolution (LR) image that can be accurately reconstructed to its original high-resolution (HR) counterpart, providing an efficient image processing and storage method for ultra-high definition media. However, extreme downscaling factors pose significant challenges to the upscaling process due to its highly ill-posed nature, causing existing image rescaling methods to struggle in generating semantically correct structures and perceptual friendly textures. In this work, we propose a novel framework called Timestep-Aware Diffusion Model (TADM) for extreme image rescaling, which performs rescaling operations in the latent space of a pre-trained autoencoder and effectively leverages powerful natural image priors learned by a pre-trained text-to-image diffusion model. Specifically, TADM adopts a pseudo-invertible module to establish the bidirectional mapping between the latent features of the HR image and the target-sized LR image. Then, the rescaled latent features are enhanced by a pre-trained diffusion model to generate more faithful details. Considering the spatially non-uniform degradation caused by the rescaling operation, we propose a novel time-step alignment strategy, which can adaptively allocate the generative capacity of the diffusion model based on the quality of the reconstructed latent features. Extensive experiments demonstrate the superiority of TADM over previous methods in both quantitative and qualitative evaluations. The code will be available at: \url{https://github.com/xxx/xxx}.
\end{abstract}
\section{Introduction}
\label{sec:intro}

With the explosive growth of ultra high-resolution (HR) images, image rescaling has become essential for enhancing data storage and transmission efficiency \cite{yang2023self}, optimizing resource usage in real-time applications \cite{qi2023real}, and ensuring proper display on devices with varying resolutions \cite{pan2022towards}. During image rescaling, HR images are downscaled to low-resolution (LR) images for storage, transmission, or display. When HR images are needed, an upscaling process is employed to restore the downscaled images to their original resolution. Unlike image super-resolution (SR), image rescaling allows for joint optimization of both downscaling and upscaling processes, preserving detailed information crucial for effective upscaling.

Recently proposed image rescaling methods can be classified into two main categories: encoder-decoder architecture \cite{kim2018task, sun2020learned} and invertible neural networks \cite{xiao2023invertible, liang2021hierarchical}. The encoder-decoder architecture employs two neural networks to perform downscaling and upscaling. While, invertible neural networks view image rescaling as a reversible process, using the forward and backward passes of an invertible neural network to simulate the downscaling and upscaling. Despite the significant improvements in reconstruction quality achieved by these methods, they are limited to small rescaling factors, such as 2$\times$ or 4$\times$.

However, with the advancement of ultra HR imaging devices, there is an increasing need for extreme rescaling factors (\textit{e.g.}, 16$\times$ or 32$\times$) to achieve more efficient storage and transmission of ultra HR data \cite{zhong2022faithful, jia2024generative}. Existing image rescaling methods often struggle to generate sufficient detail under extreme rescaling factors. Therefore, some approaches propose using pre-trained GANs as priors to constrain the solution space of the reconstructed results. GRAIN \cite{zhong2022faithful} and BDFlow \cite{li2024boundary} use pre-trained StyleGAN \cite{karras2019style, karras2020analyzing} to generate high-quality upscaled face images. VQIR \cite{wei2024towards} leverages the high-quality visual embeddings encapsulated in pre-trained VQGAN \cite{esser2021taming} to achieve extreme rescaling on natural images. Although these methods have made significant progress in perceptual metrics, their reconstruction results still suffer from incorrect semantics and structure.

To address the aforementioned issues, we propose a novel framework named Timestep-Aware Diffusion Model (TADM) for extreme image rescaling, which performs image rescaling in the latent space of a pre-trained autoencoder and leverages the natural image priors stored in large text-to-image diffusion models to restore HR images. Specifically, first, the HR images are encoded by the via the VAE encoder to obtain the latent features. Then, we use an autoencoder to rescale the latent code of the HR image to the target size, and employ a set of invertible neural networks (INN) to perform the mapping from the feature space to the pixel space to obtain the LR image. Next, the rescaled features are fed into a pre-trained diffusion model to perform a single denoising step, thereby enhancing their perceptual quality. Finally, the enhanced features are decoded via the VAE decoder to obtain the rescaled image. Additionally, we find that the rescaled latent features exhibit non-uniform degradation, which can be interpreted as the addition of noise at varying levels. As a result, different images require different time-steps to perform the denoising process. Previous studies have shown that the time-step in diffusion models directly controls the generative capacity of the model \cite{wang2023not}. Therefore, we propose a time-step alignment mechanism that predicts the appropriate time-step based on the reconstruction quality of the rescaled latent features, thereby adaptively tuning the generative capacity of the diffusion model.

Our contributions can be summarized as follows:
\begin{itemize}
\item We propose a Timestep-Aware Diffusion Model that performs well in global semantic and structural reconstruction for the extreme image rescaling task.
\item We propose a novel decoupled strategy for feature rescaling. It decouples the generation of LR images from the rescaling operation, thereby improving the reconstruction accuracy of the rescaling process.
\item We design a time-step alignment strategy to address the spatially non-uniform reconstruction quality of the rescaled latent features, allowing for more effective utilization of the prior information from the text-to-image diffusion model.
\end{itemize}

\begin{figure*}[htb]
\centering
\includegraphics[width=\linewidth]{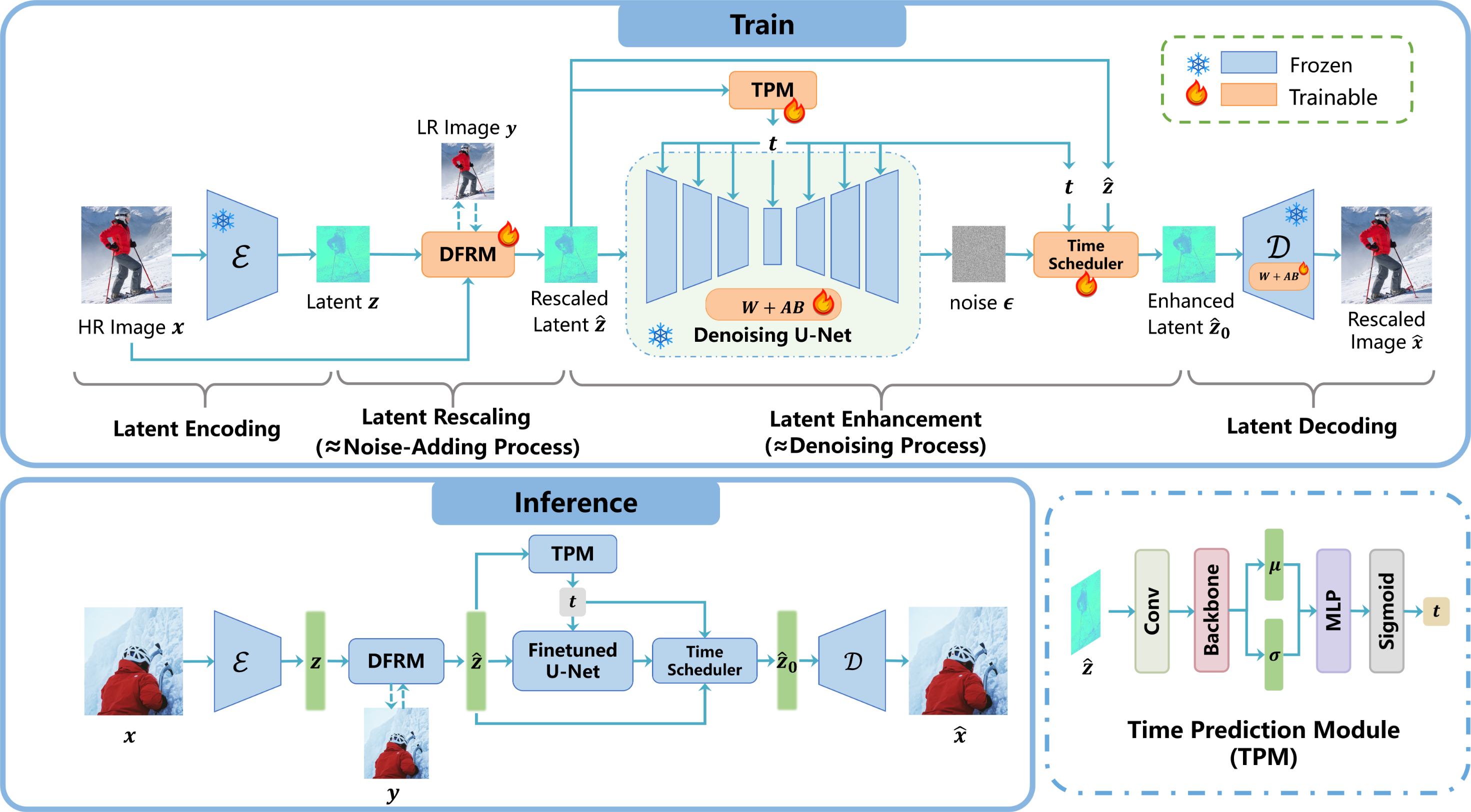}
\caption{Overview of the proposed Timestep-Aware Diffusion Model (TADM). First, the HR images $x$ are encoded to obtain the latent features $z$. The latent features are then rescaled to the target size using the decoupled feature rescaling module (DFRM), producing the LR images $y$ and outputting the rescaled latent features $\hat{z}$. Next, the rescaled latent features $\hat{z}$ are passed into a pre-trained diffusion model to perform a single denoising step for perceptual enhancement to obtain $\hat{z}_{0}$. In this process, a time prediction module (TPM) is used to estimate the time step $t$ based on $\hat{z}$ and the predicted $t$ is then fed into both the U-Net and the time scheduler. Finally, the perceptually enhanced latent features $\hat{z}_{0}$ are decoded to obtain the rescaled image $\hat{x}$.}
\label{fig:arch}
\end{figure*}

\section{Related Work}
\label{sec:Related_Work}

\subsection{Image Rescaling}
Image rescaling aims to downscale HR images to visually coherent LR images and then plausibly reconstruct the original HR images. Unlike image SR, image rescaling involves joint modeling of both downscaling and upscaling processes, leading to higher reconstruction accuracy. Earlier researches \cite{kim2018task, li2018learning} utilize CNNs to build an encoder-decoder framework, treating image rescaling as a unified task by jointly training the downscaling and upscaling processes. Instead of directly generating LR images, CAR \cite{sun2020learned} proposes generating content-adaptive resampling kernels based on HR images to perform the downsampling operation, avoiding pixel-level constraints on the downscaled images. To fully leverage the inherent reversibility of image rescaling, there has been a surge in the use of INNs for this task \cite{li2021approaching, zhu2022high, zhang2022enhancing, xu2023downscaled}. IRN \cite{xiao2023invertible} is the first invertible framework that models image rescaling as a bijective transformation, embedding residual high-frequency (HF) components into a case-agnostic latent distribution for efficient reconstruction. HCFlow \cite{liang2021hierarchical} assumes that the HF information depends on the LR image, thereby achieving better performance by incorporating LR conditions in HF information estimation. In extreme image rescaling, generative priors are usually employed to constrain the outcomes. GRAIN \cite{zhong2022faithful} and VQIR \cite{wei2024towards} leverage pre-trained StyleGAN \cite{karras2019style, karras2020analyzing} and VQGAN \cite{esser2021taming}, respectively, to achieve notable enhancements in perceptual quality. Nonetheless, these approaches are often limited in image contents, either confined to handling images from specific domains or encountering difficulties when reconstructing faces and textual elements.


\subsection{Diffusion Model for Image SR}
Here, we briefly introduce the application of diffusion models in a related task, \textit{i.e.}, image SR. Diffusion models promote the development of natural image SR in two main ways: training diffusion models from scratch \cite{li2022srdiff, gao2023implicit, yue2023resshift} and leveraging pre-trained SD model \cite{lin2024diffbir, yang2024pixel, wu2024seesr}. The former is represented by SR3 \cite{saharia2022image}, which achieves perceptual quality comparable to GANs. The latter is represented by StableSR \cite{wang2023exploiting}, which aims to build a model capable of handling image restoration under any complex degradation in real-world scenarios.

However, the above methods typically require dozens or even hundreds of iterations during inference, which significantly increases computational costs. Researchers have begun exploring how to achieve image SR using one-step diffusion models. SinSR \cite{wang2024sinsr}, based on ResShift \cite{yue2023resshift}, derives a deterministic one-step sampling process and employs a consistency preserving loss to distill the prior knowledge from a multi-step teacher network into a one-step student network. A representative work, OSEDiff \cite{wu2024one}, directly feeds the LQ image into the SD model and employs variational score distillation for regularization. S3Diff \cite{zhang2024degradation} proposes a degradation-aware LoRA and utilizes a pre-trained DINO \cite{zhang2022dino} model as a discriminator for adversarial training. InvSR \cite{yue2024arbitrary} designs a partial noise prediction strategy to provide a dynamic starting sampling point and supports inference with an arbitrary number of sampling steps ranging from one to five.


\section{Methodology}

\subsection{Overview of TADM}
As illustrated in Fig. \ref{fig:arch}, TADM consists of four parts: \textbf{Latent Encoding}, \textbf{Feature Rescaling in the Latent Space}, \textbf{Denoising Guided Perception Enhancement} and \textbf{Latent Decoding}. First, the input HR images $x$ are mapped to the latent features $z$ via the pre-trained VAE encoder $\mathcal{E}$. Subsequently, the proposed Decoupled Feature Rescaling Module (DFRM) rescales $z$ to the target size, outputting LR images $y$ and the rescaled latent features $\hat{z}$. Then, the rescaled features $\hat{z}$ are fed into a denoising U-Net network to perform a single denoising step to get the perceptually enhanced latent features $\hat{z}_{0}$. Finally, $\hat{z}_{0}$ is decoded via the pre-trained VAE decoder $\mathcal{D}$ to obtain the rescaled image $\hat{x}$. Overall, the latent rescaling process can be regarded as a noise-adding process, while the latent enhancement process can be viewed as a denoising process. To dynamically estimate the `noise' density introduced by latent rescaling, we design a time alignment strategy, which consists of a time-step prediction module (TPM) and a time scheduler module. It can better address the non-uniform degradation caused by rescaling, thereby enabling the dynamic allocation of generative capabilities of pre-trained SD.

\subsection{Feature Rescaling in the Latent Space}
\label{subsec:feature_rescaling_latent_space}
\subsubsection{Network Architecture.}
In this part, we perform rescaling of the HR latent features $z$, obtaining LR images $y$ and rescaled features $\hat{z}$. A straightforward idea is to construct a transformation chain as $y = G_{\mathrm{e}}\left(z\right)$, $\hat{z} = G_{\mathrm{d}}\left(y\right)$. Here, $G_{\mathrm{e}}$ is the encoder used to downscale the input features to the target sized LR images, while $G_{\mathrm{d}}$ represents the decoder used to reconstruct the original HR features. Guidance loss $\mathcal{L}_{\mathrm{gui.}}$ and reconstruction loss $\mathcal{L}_{\mathrm{rec.}}$ are applied to $y$ and $\hat{z}$ respectively, to force $y$ to have the same content as $x$, and $\hat{z}$ to be as similar as possible to $z$:
\begin{gather}
    \mathcal{L}_{\mathrm{gui.}} = \parallel y-\mathrm{Bicubic}(x) \parallel_{1}
    \label{equ:r2_gui_loss} \\
     \mathcal{L}_{\mathrm{rec.}} = \parallel \hat{z}-z \parallel _{1}
    \label{equ:r2_rec_loss}
\end{gather}
However, these two losses are contradictory to each other. When the guidance loss approaches zero, the rescaling problem degenerates into a classical SR problem \cite{wang2020deep}, embedding little information about the downscaling process into the LR image. In contrast, when the reconstruction loss reaches its global optimum, the rescaling problem degenerates into an image compression problem \cite{ma2019image}, causing the LR image to lack meaningful visual content. Therefore, directly applying the guidance loss $\mathcal{L}_{\mathrm{gui.}}$ on $y$ could adversely impact subsequent upscaling.

\begin{figure}[tb]
\centering
\includegraphics[width=\linewidth]{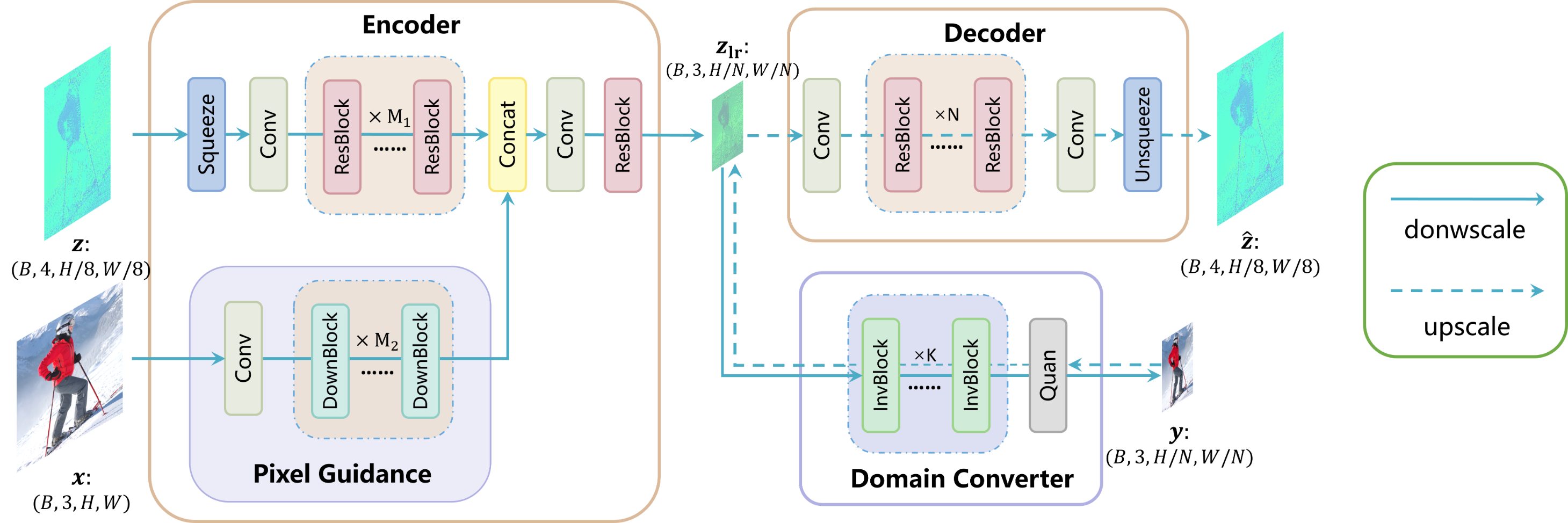}
\caption{Detailed architecture of Decoupled Feature Rescaling Module (DFRM).}
\label{fig:arch_dfrm}
\end{figure}

Considering the above issues, we devise the DFRM as shown in Fig. \ref{fig:arch_dfrm}, which divides the rescaling operation into two sub-tasks: feature rescaling and LR image generation. Inspired by FGRN \cite{li2021approaching}, we use a set of CNNs to perform downscaling and upscaling, and a separate INN to perform the bidirectional mapping between the feature domain and the pixel domain. To achieve a balance between feature rescaling accuracy and visual quality of LR images, we introduce another pixel guidance module in the encoder to incorporate pixel-level information, which can improve the quality of the LR image. Specifically, DFRM perform feature rescaling through two independent transformation chains $\left(\left(x, z\right) \rightarrow z_\text{lr} \rightarrow \hat{z}\right)$; $\left(z_\text{lr} \leftrightarrow y\right)$ as follows:
\begin{gather}
     z_\text{lr} = G_{\mathrm{e}}\left(x, z\right)
    \label{equ:r2_my_encoder} \\
    z_\text{lr} \approx  F^{-1}\left(F\left(z_\text{lr}\right)\right) = F^{-1}\left(y\right)
    \label{equ:r2_my_inn} \\
     \hat{z} = G_{\mathrm{d}}\left(z_\text{lr}\right)
    \label{equ:r2_my_decoder}
\end{gather}
Here, $F$ and $F^{-1}$ represent the forward and backward processes of the INN, respectively. Given the latent variable $z \in \mathbb{R}^{\frac{H}{8} \times \frac{W}{8} \times 4}$ of the pre-trained SD VAE as input, the encoder $G_{\mathrm{e}}$ first transforms it into a compact representation $z_\text{lr} \in \mathbb{R}^{\frac{H}{N} \times \frac{W}{N} \times 3}$, whose size is the same as that of the target LR image and $N$ represents the target downscaling factor. Subsequently, a set of INNs perform the forward process to transform $z_\text{lr}$ from the feature domain to the pixel domain and obtain the LR image $y \in \mathbb{R}^{\frac{H}{N} \times \frac{W}{N} \times 3}$ by quantization. In the upscaling process, the inverse of the INN is used to process the LR image $y$ and reconstruct $z_\text{lr}$. Finally, the decoder $G_{\mathrm{d}}$ is utilized to obtain the rescaled feature $\hat{z}$.

\subsubsection{Loss Function.}
Training of this part involves two losses: reconstruction loss and guidance loss. Considering the information lost in quantization operations, the bidirectional mapping between $z_\text{lr}$ and $y$ is not exactly reversible. To ensure the robustness of the network to quantization operations, we consider two types of transformation chains in designing the reconstruction loss: $\left(x, z\right) \rightarrow z_\text{lr} \rightarrow \hat{z}$ and $\left(x, z\right) \rightarrow z_\text{lr} \rightarrow y \rightarrow \hat{z}_{\text{lr}} \rightarrow \hat{z}$, which can be expressed as the following formula:
\begin{equation}
    \begin{split}
         \mathcal{L}_{\mathrm{rec.}} = & \parallel G_{\mathrm{d}}\left(G_{\mathrm{e}}\left(x, z\right)\right) -z \parallel_{1} + \\
         & \parallel G_{\mathrm{d}}\left(F^{-1}\left(F\left(G_{\mathrm{e}}\left(x, z\right)\right)\right)\right) -z \parallel_{1}
        \label{equ:r2_my_rec_loss}
    \end{split}
\end{equation}
This combination can facilitate the network to reconstruct $\hat{z}$ with high fidelity in both the existence and absence of quantization operations. Additionally,  the guidance loss is designed as follows:
\begin{equation}
    \mathcal{L}_{\mathrm{gui.}} = \parallel F\left(G_{\mathrm{e}}\left(x, z\right)\right) -\mathrm{Bicubic}(x) \parallel_{1}
    \label{equ:r2_my_gui_loss} \\
\end{equation}
Finally, the loss function of this part is a combination of these two losses:
\begin{equation}
    \mathcal{L}_{\mathrm{res.}} = \lambda_\mathrm{rec.} \mathcal{L}_{\mathrm{rec.}} + \lambda_\mathrm{gui.} \mathcal{L}_{\mathrm{gui.}}
    \label{equ:r2_stage1_loss} \\
\end{equation}

\subsection{Denoising Guided Perception Enhancement}
\label{subsec:one_step_perception_enhancement}
The latent variable $\hat{z}$ obtained in Sec. \ref{subsec:feature_rescaling_latent_space} retains sufficient semantic information about the HR image, but lacks adequate texture. Therefore, we propose to leverage the strong priors stored in the pre-trained SD model to enhance the perceptual quality of the rescaling results in this part. Original diffusion model requires a multi-step sampling algorithm, significantly increases inference time and makes it difficult to scale to practical applications. Inspired by recent work on accelerating diffusion models \cite{kang2024distilling}, we adopt an one-step inference strategy. Specifically, we can directly feed the rescaled latent variable $\hat{z}$ into the denoising U-Net $\epsilon = \epsilon_{\theta}\left(\hat{z}, t\right)$ and perform a single denoising step as follows:
\begin{gather}
    \hat{z}_{0} = \frac{\hat{z}-\sqrt{1-\bar{\alpha}_{t}}\epsilon}{\sqrt{\bar{\alpha}_{t}}}
    \label{equ:denoise}
\end{gather}
where $\hat{z}_{0}$ represents the perceptually enhanced latent features, $\bar{\alpha}_{t}$ is the noise scheduler of the forward diffusion process of the SD model, and $t$ represents the time-step associated with the noise intensity. Finally, $\hat{z}_{0}$ can be transformed to the pixel domain through the pre-trained VAE decoder $\mathcal{D}$ to obtain the upscaled image $\hat{x} = \mathcal{D}\left(\hat{z}_{0}\right)$. Then, we can adopt the loss function between $\hat{x}$ and $x$ to optimize the U-Net using LoRA layers.

\begin{figure}[tb]
\centering
\includegraphics[width=\linewidth]{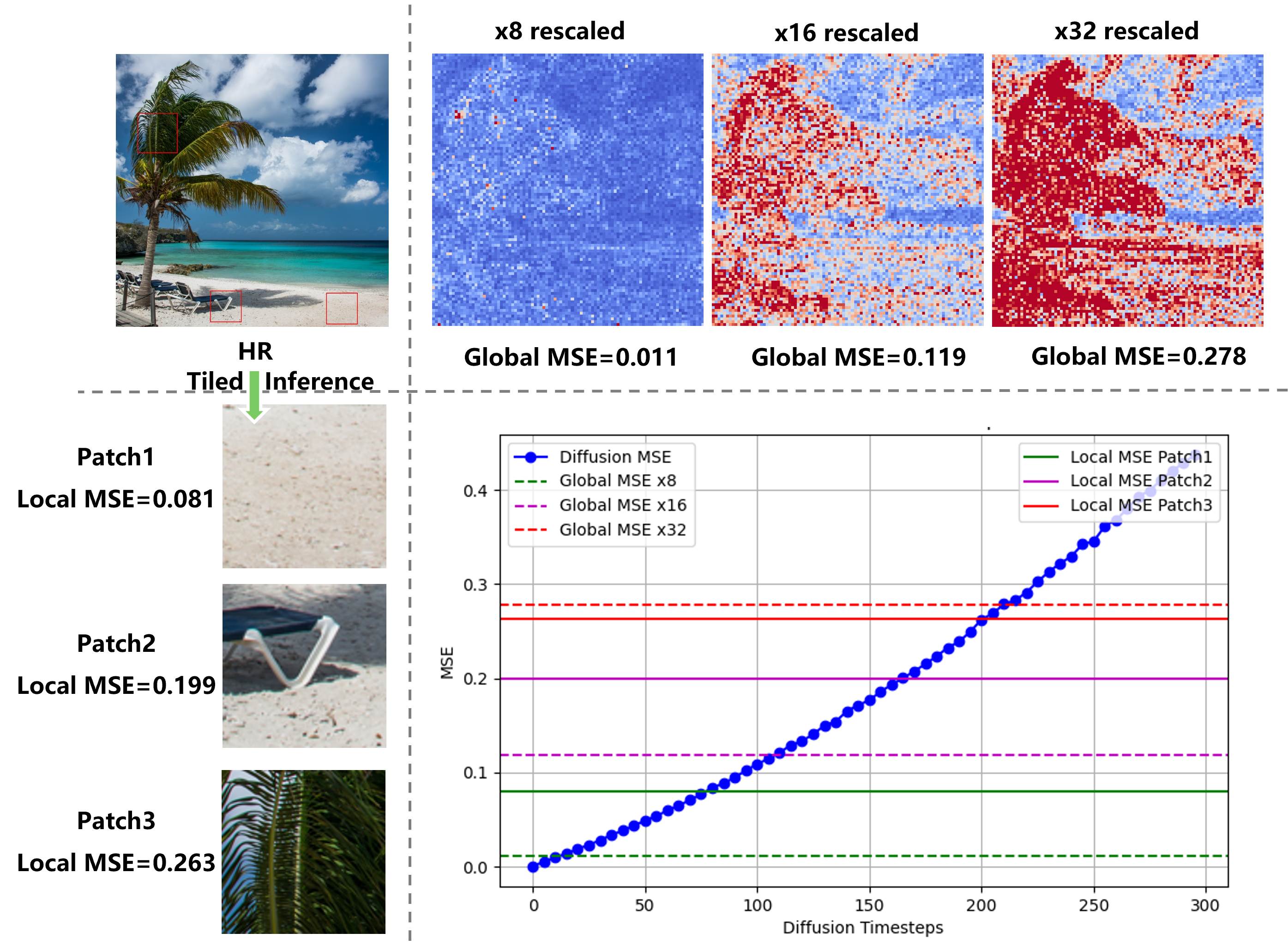}
\caption{We compare the MSE introduced by the rescaling operation with that caused by the forward diffusion process. The rescaling degradation severity is correlated with both the image content and the rescaling scale. Therefore, the model requires adaptive time-step to align the rescaling MSE with the diffusion MSE.}
\label{fig:motivation}
\end{figure}

However, the key problem of the aforementioned approaches lies in the selection of time-step $t$. We can regard the rescaling process as the noise-adding process of diffusion model, while Eq. (\ref{equ:denoise}) represents the denoising process. For different rescaling factors and different images, the `noise' introduced by the rescaling operation varies in intensity. In Fig. \ref{fig:motivation}, we present the global rescaling error of an image under different rescaling factors, as well as the local rescaling error of different patches within the image. Additionally, we plot the error curve of the forward noise-adding process of the SD model. It can be observed that as the rescaling factor and image content change, rescaling MSE also varies and corresponds to the diffusion MSE at different time-steps. Thus, we propose to dynamically predict the time-step $t$ based on the quality of the rescaled latent features $\hat{z}$. A simple yet effective network named time-step prediction module (TPM), whose structure is presented in Fig. \ref{fig:arch}, is used to predict $t$ based on the rescaled latent features: $t = \mathrm{TPM}\left(\hat{z}\right)$. During the training phase, we compute a single time-step for each input image patch. In the inference phase, we adopt a tiled inference approach (which will be elaborated in detail in the supplementary materials), allowing for predicting different time-steps for various regions of a high-resolution input image.

\begin{figure}[tb]
\centering
\includegraphics[width=\linewidth]{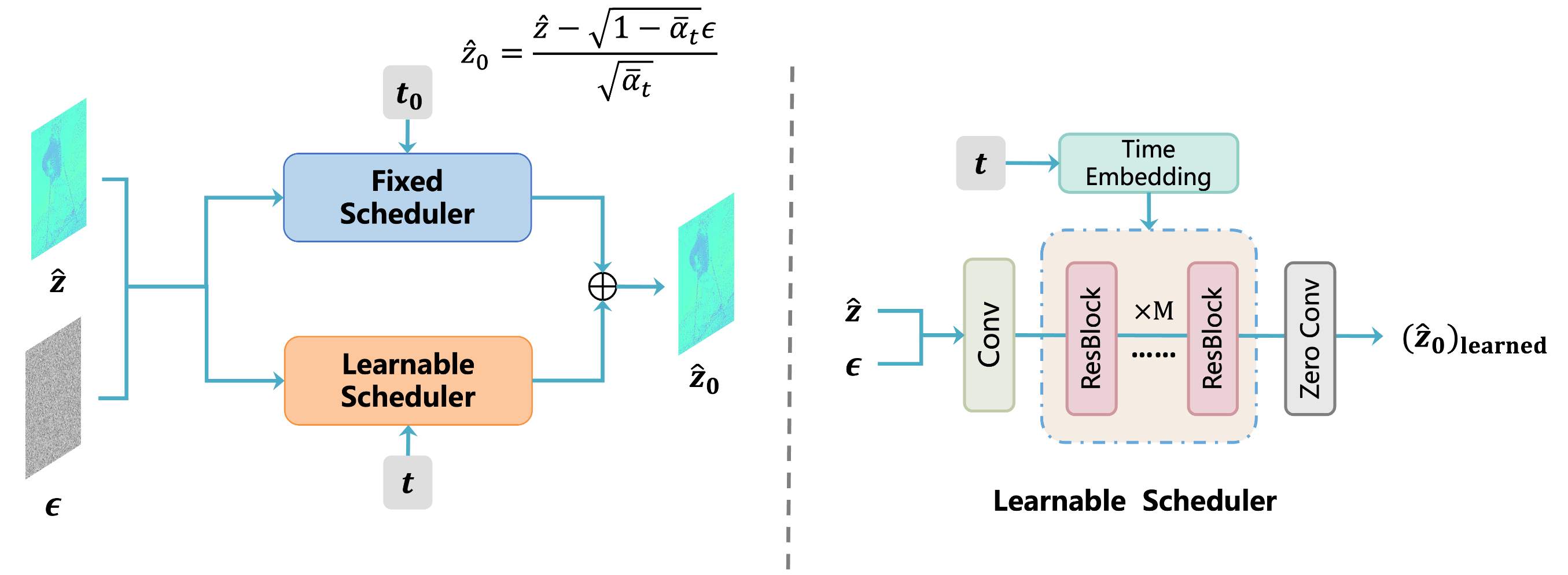}
\caption{The structure of the time scheduler module. It combines both the fixed scheduler and learnable scheduler to perform the denoising process. The two schedulers are connected by a zero-convolution layer to stabilize the training.}
\label{fig:refiner}
\end{figure}

Subsequently, the predicted time-step $t$ is used to perform a single denoising step. However, the vanilla SD time scheduler as shown in Eq. (\ref{equ:denoise}) is not differentiable with respect to $t$, which prevents gradients from being back-propagated to the TPM. Considering that given $\hat{z}$, $\epsilon$ and $t$, Eq. (\ref{equ:denoise})  essentially represents a linear transformation, which can be learned by a neural network. Therefore, we propose a hybrid time scheduler module consisting of a fixed scheduler and a learnable scheduler, whose structure is presented in Fig. \ref{fig:refiner}. The former uses Eq. (\ref{equ:denoise}) to perform the denoising step, while the latter employs a neural network to simulate the denoising scheduler. To ensure training stability, the final layer of the learnable scheduler is a zero-initialized convolution, allowing it to gradually correct the results of the fixed scheduler as training progresses. This process can be described as follows:
\begin{equation}
    \begin{split}
         \hat{z}_{0} = \mathcal{S}_\mathrm{fixed}\left(\hat{z}, \epsilon, t_{0}\right) + \mathcal{S}_\mathrm{learned}\left(\hat{z}, \epsilon, t\right)
    \label{equ:refiner}
    \end{split}
\end{equation}
where $t_{0}$ is the fixed pre-set time-step and $\mathcal{S}_\mathrm{fixed}$ denotes the vanilla SD time scheduler as shown in Eq. (\ref{equ:denoise}).

\begin{table*}[htb]
    \caption{Quantitative comparisons with different methods at 16$\times$ and 32$\times$. The symbols $\uparrow$ and $\downarrow$ respectively represent that higher or lower values indicate better performance. Bold represents the best and underline represents the second best.}
    \label{tab:quantitative_comparison}
    \centering
    \renewcommand\arraystretch{1}
    \setlength{\tabcolsep}{1.75mm}
    {\small\begin{tabularx}{\textwidth}{c|c|cc|cc|cc|cc|cc|cc}
    \hline
    \multirow{2}{*}{Dataset} & \multirow{2}{*}{Method} & \multicolumn{2}{c|}{PSNR $\uparrow$} & \multicolumn{2}{c|}{SSIM $\uparrow$} & \multicolumn{2}{c|}{LPIPS $\downarrow$} & \multicolumn{2}{c|}{DISTS $\downarrow$} & \multicolumn{2}{c|}{MUSIQ $\uparrow$} & \multicolumn{2}{c}{CLIPIQA $\uparrow$} \\
    \cline{3-14}
    & & 16$\times$ & 32$\times$ & 16$\times$ & 32$\times$ & 16$\times$ & 32$\times$ & 16$\times$ & 32$\times$ & 16$\times$ & 32$\times$ & 16$\times$ & 32$\times$ \\
    \hline
    
    \multirow{7}{*}{DIV2K}
    & ESRGAN & 23.15 & 19.94 & 0.5946 & 0.5689 & 0.4478 & 0.5892 & 0.2378 & 0.4230 & 59.80 & 43.35 & 0.6161 & 0.3734 \\
    & StableSR & 22.02 & 19.92 & 0.5983 & 0.5588 & 0.4980 & 0.5736 & 0.2078 & 0.2603 & 51.28 & 48.94 & 0.3780 & 0.3203 \\
    & S3Diff & 20.22 & 17.81 & 0.5491 & 0.4601 & 0.4033 & 0.4895 & 0.1309 & \underline{0.1810} & \underline{64.37} & \underline{67.92} & 0.6228 & \underline{0.6991} \\
    & IRN & \underline{26.04} & \underline{22.84} & \underline{0.7020} & \underline{0.6212} & 0.5153 & 0.6095 & 0.3041 & 0.4036 & 43.72 & 28.70 & 0.2799 & 0.2336 \\
    & HCFlow & \textbf{26.66} & \textbf{23.89} & \textbf{0.7176} & \textbf{0.6467} & 0.4885 & 0.5816 & 0.2866 & 0.3852 & 46.43 & 37.25 & 0.2735 & 0.2792 \\
    & VQIR & 23.91 & 22.02 & 0.6498 & 0.5823 & \underline{0.3174} & \underline{0.4568} & \underline{0.1024} & 0.2663 & 64.04 & 58.21 & \underline{0.6350} & 0.6293 \\
    & Ours & 23.98 & 22.18 & 0.6736 & 0.5869 & \textbf{0.2979} & \textbf{0.4221} & \textbf{0.0886} & \textbf{0.1684} & \textbf{66.56} & \textbf{69.12} & \textbf{0.7189} & \textbf{0.7204} \\
    \hline
    
    \multirow{7}{*}{CLIC2020}
    & ESRGAN & 25.43 & 22.47 & 0.7018 & 0.6804 & 0.4213 & 0.5429 & 0.2209 & 0.4011 & 57.80 & 41.28 & \underline{0.6107} & 0.3558 \\
    & StableSR & 24.11 & 22.24 & 0.7052 & 0.6726 & 0.4625 & 0.5288 & 0.1860 & 0.2282 & 52.69 & 49.80 & 0.3779 & 0.3325 \\
    & S3Diff & 22.04 & 19.87 & 0.6615 & 0.5711 & 0.3968 & 0.4829 & 0.1336 & \underline{0.1812} & 63.07 & \underline{66.40} & 0.6001 & \underline{0.6838} \\
    & IRN & \underline{28.55} & \underline{24.93} & \underline{0.7907} & \underline{0.7228} & 0.4676 & 0.5528 & 0.2746 & 0.3776 & 47.66 & 27.81 & 0.2950 & 0.2495 \\
    & HCFlow & \textbf{28.82} & \textbf{26.21} & \textbf{0.7949} & \textbf{0.7451} & 0.4542 & 0.5325 & 0.2635 & 0.3611 & 47.72 & 37.15 & 0.2733 & 0.2884 \\
    & VQIR & 25.69 & 24.12 & 0.7363 & 0.6942 & \underline{0.2854} & \underline{0.4191} & \underline{0.0901} & 0.2445 & \underline{64.36} & 54.83 & 0.6105 & 0.6068 \\
    & Ours & 26.04 & 24.26 & 0.7542 & 0.6903 & \textbf{0.2724} & \textbf{0.3911} & \textbf{0.0792} & \textbf{0.1513} & \textbf{65.35} & \textbf{69.10} & \textbf{0.6815} & \textbf{0.7271} \\
    \hline

    \multirow{7}{*}{Urban100}
    & ESRGAN & 19.36 & 17.68 & 0.4835 & 0.4344 & 0.4741 & 0.6142 & 0.2554 & 0.4288 & 65.82 & 57.02 & 0.5849 & 0.3860 \\
    & StableSR & 19.93 & 18.08 & 0.5020 & 0.4447 & 0.5827 & 0.6839 & 0.3701 & 0.4734 & 41.97 & 24.80 & 0.3439 & 0.4173 \\
    & S3Diff & 17.60 & 15.74 & 0.4368 & 0.3419 & 0.4204 & 0.5282 & 0.1917 & \underline{0.2573} & 71.01 & \underline{69.56} & \underline{0.6658} & \underline{0.6776} \\
    & IRN & \underline{22.15} & \underline{18.93} & 0.6155 & \underline{0.4772} & 0.5114 & 0.6567 & 0.3338 & 0.4457 & 58.73 & 32.03 & 0.3217 & 0.2210 \\
    & HCFlow & \textbf{22.59} & \textbf{19.86} & \underline{0.6335} & \textbf{0.5185} & 0.4841 & 0.6048 & 0.3125 & 0.4101 & 60.62 & 47.94 & 0.3182 & 0.2827 \\
    & VQIR & 20.27 & 18.44 & 0.5782 & 0.4427 & \underline{0.3038} & \underline{0.5045} & \underline{0.1418} & 0.3290 & \underline{71.56} & 61.64 & 0.6619 & 0.6568 \\
    & Ours & 21.46 & 18.84 & \textbf{0.6495} & 0.4685 & \textbf{0.2577} & \textbf{0.4481} & \textbf{0.1154} & \textbf{0.2386} & \textbf{72.22} & \textbf{72.88} & \textbf{0.7155} & \textbf{0.6795} \\
    \hline

    \multirow{7}{*}{DIV8K}
    & ESRGAN & 25.49 & 23.78 & 0.6247 & 0.6350 & 0.4628 & 0.5610 & 0.2518 & 0.4268 & 53.49 & 39.11 & \underline{0.6401} & 0.4078 \\
    & StableSR & 25.90 & 23.55 & 0.6602 & 0.6258 & 0.4844 & 0.5497 & 0.2134 & 0.2623 & 47.15 & 46.97 & 0.4017 & 0.3557 \\
    & S3Diff & 22.20 & 20.19 & 0.5903 & 0.5063 & 0.4163 & 0.4954 & 0.1524 & \underline{0.1997} & \textbf{59.98} & \underline{62.58} & 0.6351 & \underline{0.6901} \\
    & IRN & \underline{28.95} & \underline{25.46} & \underline{0.7402} & \underline{0.6636} & 0.4932 & 0.5870 & 0.3044 & 0.4066 & 42.61 & 27.76 & 0.3216 & 0.2741 \\
    & HCFlow & \textbf{29.22} & \textbf{26.69} & \textbf{0.7470} & \textbf{0.6857} & 0.4793 & 0.5616 & 0.2932 & 0.3906 & 42.71 & 35.53 & 0.2948 & 0.3128 \\
    & VQIR & 25.79 & 24.26 & 0.6889 & 0.6350 & \underline{0.3221} & \underline{0.4457} & \underline{0.1066} & 0.2628 & 56.76 & 51.44 & 0.5769 & 0.6203 \\
    & Ours & 26.16 & 24.37 & 0.7163 & 0.6337 & \textbf{0.3117} & \textbf{0.4358} & \textbf{0.0979} & \textbf{0.1888} & \underline{58.54} & \textbf{63.34} & \textbf{0.6836} & \textbf{0.7229} \\
    \hline
    \end{tabularx}}
\end{table*}

Finally, we fine-tune the denoising U-Net network and the VAE decoder using LoRA, while also jointly training the TPM and time scheduler module. To achieve a balance between fidelity and realism, we supervise the reconstruction results $\hat{x}$ using pixel loss and perceptual loss as follows:
\begin{equation}
\begin{aligned}
    \mathcal{L}_{\mathrm{enh.}} =& \left\| \hat{x} - x \right\|_1 +\\
    &\lambda_{\mathrm{pec.}} \left( \mathcal{L}_{\mathrm{lpips}} \left(\hat{x}, x\right) + \mathcal{L}_{\mathrm{dists}}\left(\hat{x}, x\right)\right)
\end{aligned}
    \label{equ:r2_stage2_loss}
\end{equation}

\section{Experiments}
\subsection{Experimental Setup}
\subsubsection{Datasets and Evaluation Metrics.}
 We train the proposed TADM using the DF2K dataset \cite{agustsson2017ntire, lim2017enhanced}. Considering that traditional benchmarks such as Set5, Set14, and BSD100 often have lower quality and resolution \cite{gu2019div8k}, they are not suitable for evaluating extreme rescaling methods. Therefore, we utilize four high-resolution datasets: DIV2K, CLIC2020 \cite{toderici2020clic}, Urban100 \cite{huang2015single}, and DIV8K \cite{gu2019div8k} to evaluate models been compared. We adopt PSNR and SSIM \cite{wang2004image} for quantitative evaluation. For perceptual quality evaluation, reference-based metrics LPIPS \cite{zhang2018unreasonable} and DISTS \cite{ding2020image} are used. For non-reference image quality assessment, we use metrics including MUSIQ \cite{ke2021musiq} and CLIPIQA \cite{wang2023exploring}.

\subsubsection{Training Details}
TADM is built based on the SD 2.1-base model, and the training is divided into three stages. First, we train the DFRM using Eq. (\ref{equ:r2_stage1_loss}). Then we adopt Eq. (\ref{equ:r2_stage2_loss}) to jointly train the LoRA layers, the TPM and the time scheduler module. Finally, we use a smaller learning rate to jointly fine-tune DFRM, LoRA layers, TPM and scheduler module using a combination of Eq. (\ref{equ:r2_stage1_loss}) and (\ref{equ:r2_stage2_loss}). More training details are provided in the supplementary material.

\subsection{Results and Comparison with SOTA}
We compare the proposed TADM with three kinds of methods: (1) bicubic downscaling and SR, \ie, ESRGAN \cite{wang2018esrgan}, StableSR \cite{wang2023exploiting} (based on multi-step diffusion) and S3Diff (based on single-step diffusion)\cite{zhang2024degradation}; (2) flow-based rescaling models, \ie, IRN \cite{xiao2023invertible} and HCFlow \cite{liang2021hierarchical}; (3) prior-based extreme rescaling method VQIR \cite{wei2024towards}. Some of them are proposed for 4$\times$ rescaling factors and we extended them to 16$\times$ or 32$\times$. To ensure a fair comparison, all methods are trained until convergence on the DF2K dataset following their original configurations.

\subsubsection{Quantitative Results}
Table \ref{tab:quantitative_comparison} presents the 16$\times$ and 32$\times$ quantitative image rescaling results on four datasets. The performance of SR methods are generally poor. Both gan-based (ESRGAN) and diffusion-based (StableSR and S3Diff) methods can only achieve relative good performance on no-reference metrics. This is because they rely solely on the LR image to reconstruct the original image and cannot jointly optimize the downscaling and upscaling processes. While methods based on INNs (HCFlow and IRN) achieve the highest PSNR and SSIM due to its consideration of the correlation between downscaling and upscaling. In all generative methods, VQIR achieves relatively good perceptual metrics by leveraging the priors of the pre-trained VQGAN. In contrast, our method achieves optimal performance across all perceptual metrics on all datasets. For instance, in the 32$\times$ rescaling on the DIV2K validation set, our method improves the DISTS metric by 36.76$\%$ compared to the second-best VQIR. Moreover, in the 16$\times$ rescaling on the Urban100 dataset, SSIM of our method even surpasses that of the INN-based regression model, reaching optimal performance. This demonstrates the advantage of our method in reconstructing the structure of images.

\begin{figure*}[htb]
\centering
\includegraphics[width=0.95\linewidth]{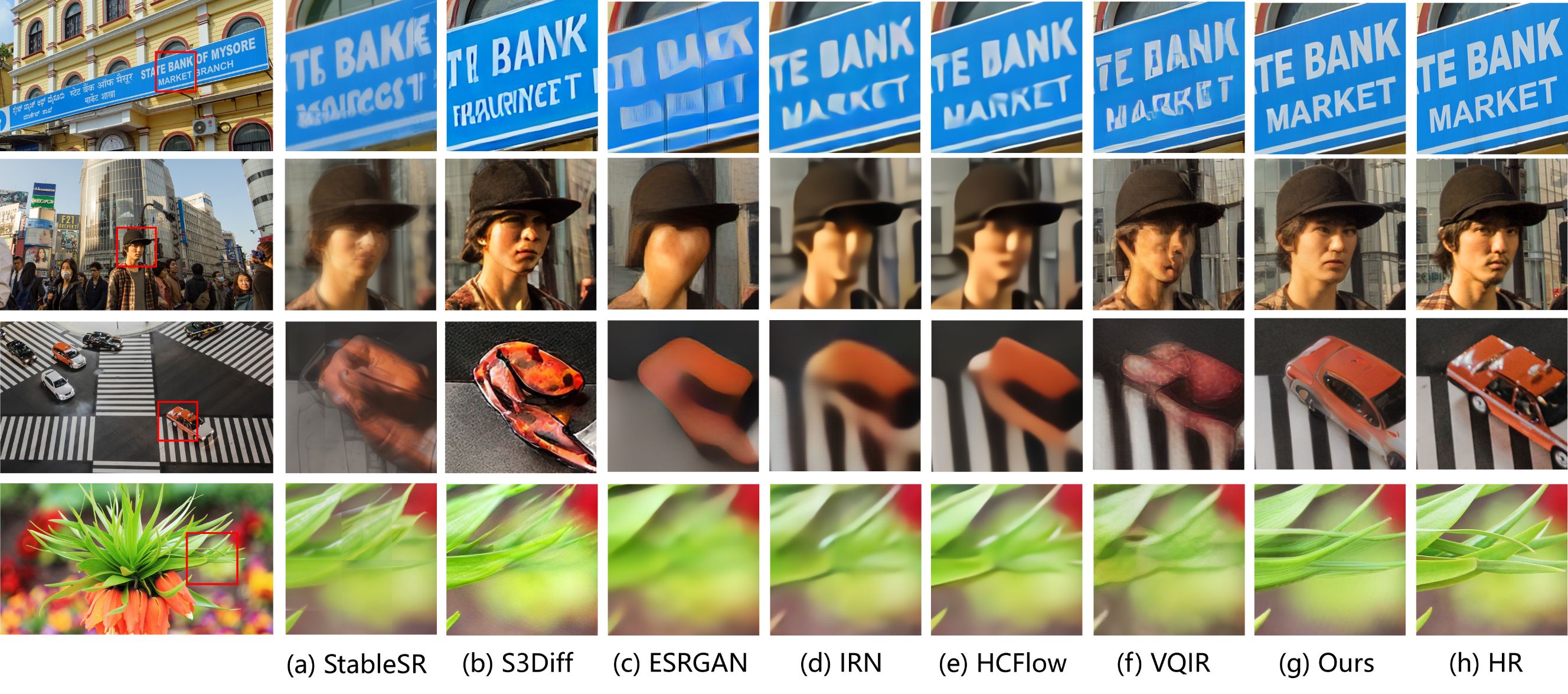}
\caption{Qualitative comparisons of 16$\times$ (the 1st and 2nd rows) and 32$\times$ (the 3rd and 4th rows) rescaling results. Our method achieves higher semantic accuracy, such as more clearer text, more recognizable faces, and more realistic structures.}
\label{fig:qualitative_compare}
\end{figure*}

\begin{figure*}[htb]
    \centering
    \includegraphics[width=\linewidth]{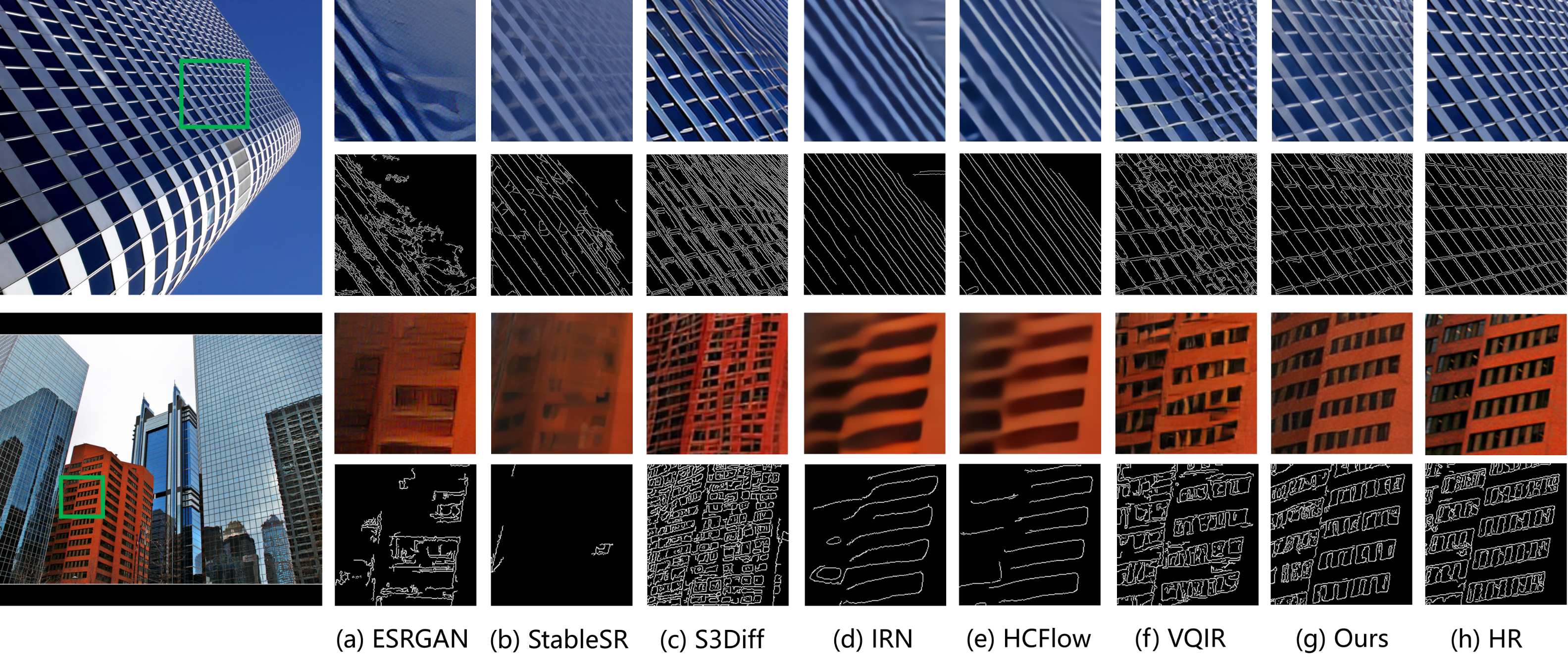}
    \caption{Visual comparisons of 16$\times$ rescaling methods on the Urban100 dataset, including both the rescaled images and corresponding edge maps. Our model is capable of restoring more regular structures and producing sharper, more accurate edges.}
    \label{fig:more_compare_urban100}
\end{figure*}

\begin{figure*}[htb]
    \centering
    \includegraphics[width=\linewidth]{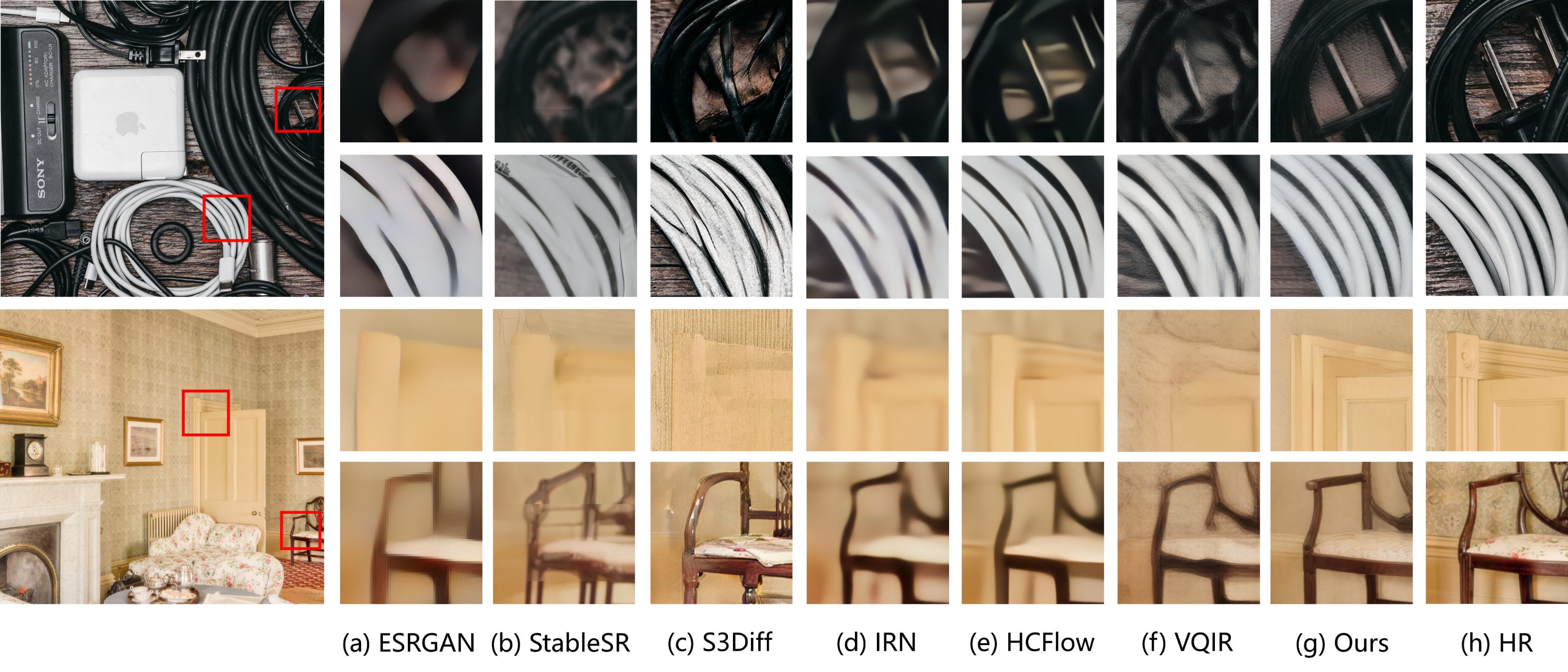}
    \caption{Visual comparisons of 32$\times$ rescaling methods on the DIV8K dataset. Our approach can still reconstruct correct semantic information even in such extreme scenarios.}
    \label{fig:more_compare_div8k}
\end{figure*}

\subsubsection{Qualitative Results}
In Fig. \ref{fig:qualitative_compare}, we present visual comparisons at 16$\times$ and 32$\times$ rescaling factors. Due to the ill-posed nature of extreme rescaling, pixel-oriented optimization methods such as IRN, and HCFlow produce overly-smoothed results. Although ESRGAN incorporates perceptual optimization it still cannot generate adequate details. VQIR leverages the pre-trained VQGAN as a prior, allowing it to generate rich details and sharp edges. However, it struggles with faces and more complex structures. Although S3Diff can generate sharp edges and rich details, its reconstructed results have low fidelity and exhibit significant artifacts and color distortions. Thanks to the powerful prior provided by the pre-trained SD and the carefully designed architecture,, our method effectively addresses these issues. It not only generates sufficiently rich textures but also reconstructs accuracy semantics and regular structures, such as text, faces, vehicles, and leaves.

The quantitative experiments demonstrate that our model exhibits superior performance on the Urban100 \cite{huang2015single} dataset, confirming that our method produces more accurate structural features. In this section, we provide a visual comparison of 16$\times$ rescaling results on the Urban100 dataset, as shown in Fig. \ref{fig:more_compare_urban100}.  Traditional regression models, such as IRN \cite{xiao2023invertible} and HCFlow \cite{liang2021hierarchical}, tend to generate overly-smooth results. GAN-based methods, including ESRGAN \cite{wang2018esrgan} and VQIR \cite{wei2024towards}, generate more details, but their edge information lacks accuracy and regularity. For diffusion-based super-resolution methods, such as S3Diff \cite{zhang2024degradation}, they are capable of generating sharp edges. However, due to the lack of information about downscaling, their results often exhibit lower fidelity, which manifests in distorted structures. In contrast, our approach, by leveraging the prior knowledge encapsulated in the SD model, is able to reconstruct more accurate and rich edge details. In Fig. \ref{fig:more_compare_div8k}, we present a visual comparison of 32$\times$ rescaling results on the ultra-high-resolution dataset, DIV8K. It is evident that our method is capable of recovering images with accurate semantics, even at extreme rescaling factors. For instance, our method is capable of accurately reconstructing electronic devices, doors, and chairs, while retaining rich details.

\begin{figure*}[tb]
    \centering
    \includegraphics[width=0.9\linewidth]{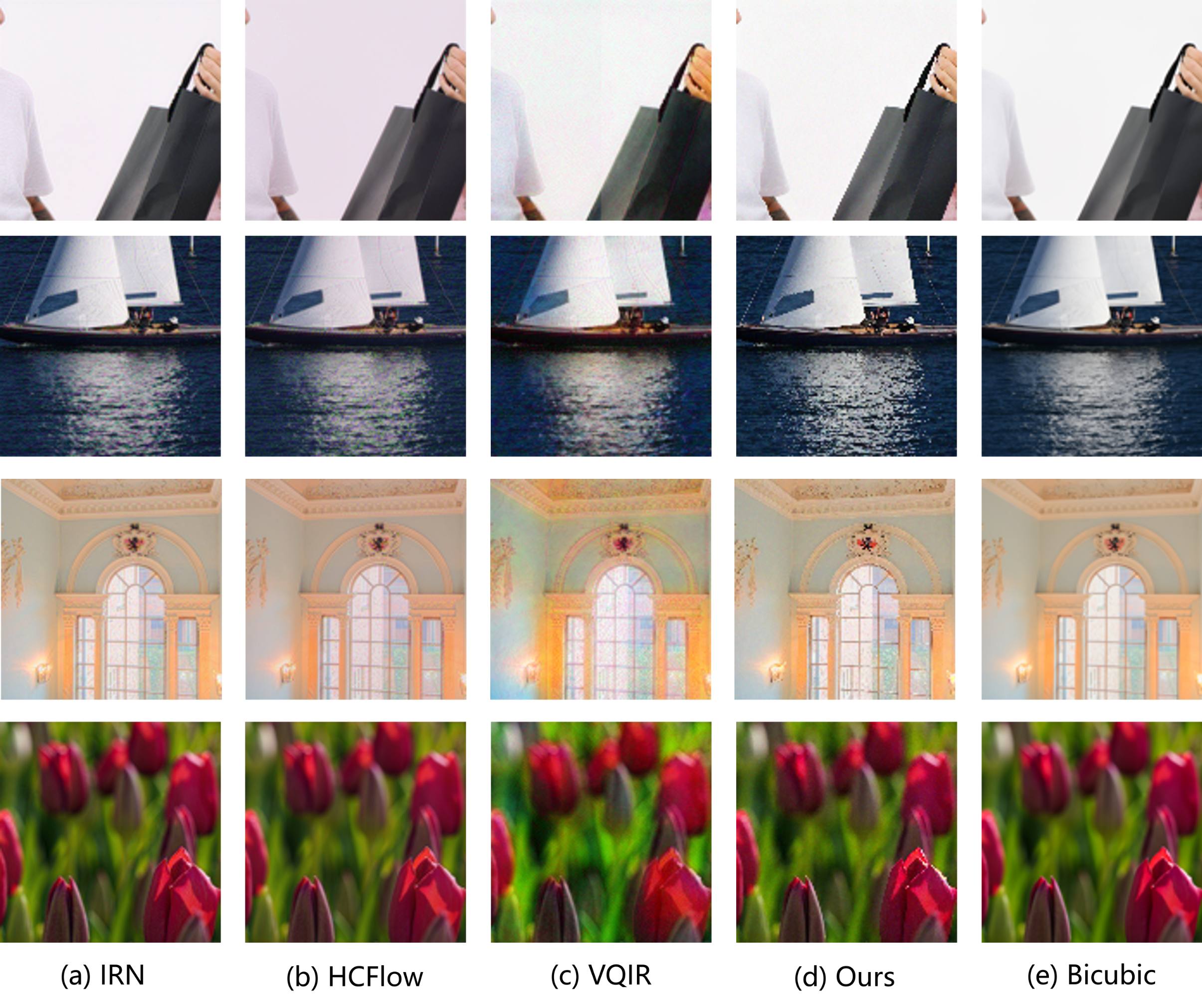}
    \caption{Visual comparisons of downscaled LR images with sota methods.}
    \label{fig:compare_lr}
\end{figure*}

To validate the effectiveness of our downscaling scheme, we conduct a visual comparison of the downscaled LR images with state-of-the-art (SOTA) methods, as shown in Fig. \ref{fig:compare_lr}. Overall, the LR images generated by our method achieve comparable or even superior visual quality to SOTA image rescaling methods. The LR images produced by VQIR exhibit noticeable color shifts. Furthermore, VQIR generated LR images contain significant noise, whereas our method exhibits a certain degree of ringing artifacts. In future work, we plan to further optimize the quality of the LR images.

\subsection{Ablation Study}
We conduct ablation experiments to validate the effectiveness of each component. All ablations are trained on the DF2K dataset and tested on the validation set of the DIV2K for 16$\times$ rescaling task.

\begin{table}[htb]
	\caption{Ablation study of rescaling space and SD prior.}
	\label{tab:ablation_study_lsbir}
	\centering
        \setlength{\tabcolsep}{1.2mm}
	\begin{tabular}{ccccc}
		\hline
         Latent space & Pixel space & SD prior & LPIPS$\ \downarrow$ & DISTS$\ \downarrow$ \\
        \hline
        $\times$ & $\checkmark$ & $\checkmark$ & 0.3630 & 0.1154 \\
        $\checkmark$ & $\times$ & $\times$ & 0.4675 & 0.3109 \\
        $\checkmark$ & $\times$ & $\checkmark$ & \textbf{0.2979} & \textbf{0.0886} \\
        \hline
	\end{tabular}
\end{table}

\subsubsection{Effectiveness of Feature Rescaling}
\begin{figure}[htb]
    \centering
    \includegraphics[width=\linewidth]{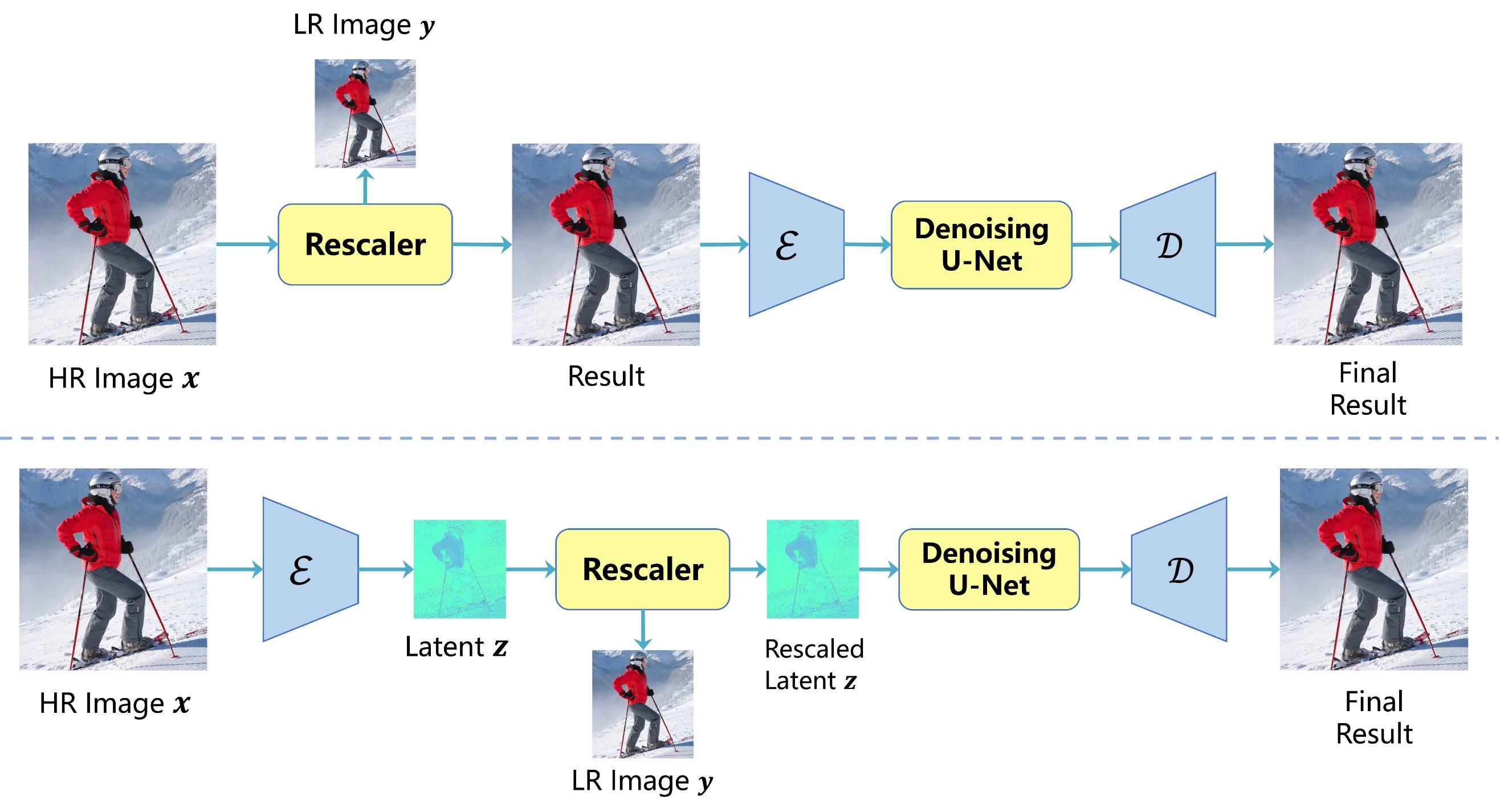}
    \caption{Image rescaling within pixel space and latent space with Stable Diffusion (SD) prior.}
    \label{fig:pixel_vs_latent}
\end{figure}

To verify the effectiveness of the feature rescaling operation, we move the rescaling operation from the latent space to the pixel space. Specifically, we use HCFlow \cite{liang2021hierarchical}, the comparative method with the best PSNR performance, to perform the rescaling operation in the pixel domain, followed by perceptual enhancement using the same approach described in Sec. \ref{subsec:one_step_perception_enhancement}. As shown in Fig. \ref{fig:pixel_vs_latent}, the rescaling operator in pixel space takes HR image as input and simultaneously outputs both the LR image and the rescaled image. Then, to use SD for perceptual enhancement, we need to map the rescaled image to the latent space using a VAE encoder and perform the denoising process. Finally, the enhanced latent features are mapped back to the pixel space using the VAE decoder. Due to the nonlinear mapping nature of the VAE encoder, the minimum distance in pixel space does not correspond to the minimum distance in latent space. This causes the rescaling operator trained in pixel space to misalign with the prior of the pre-trained SD model, resulting in a degradation of perceptual quality. As shown the 1st row in Table \ref{tab:ablation_study_lsbir}, the rescaling operation in the latent space can improve reconstruction performance, as evidenced by much lower LPIPS and DISTS.

\begin{figure}[htb]
    \centering
    \includegraphics[width=\linewidth]{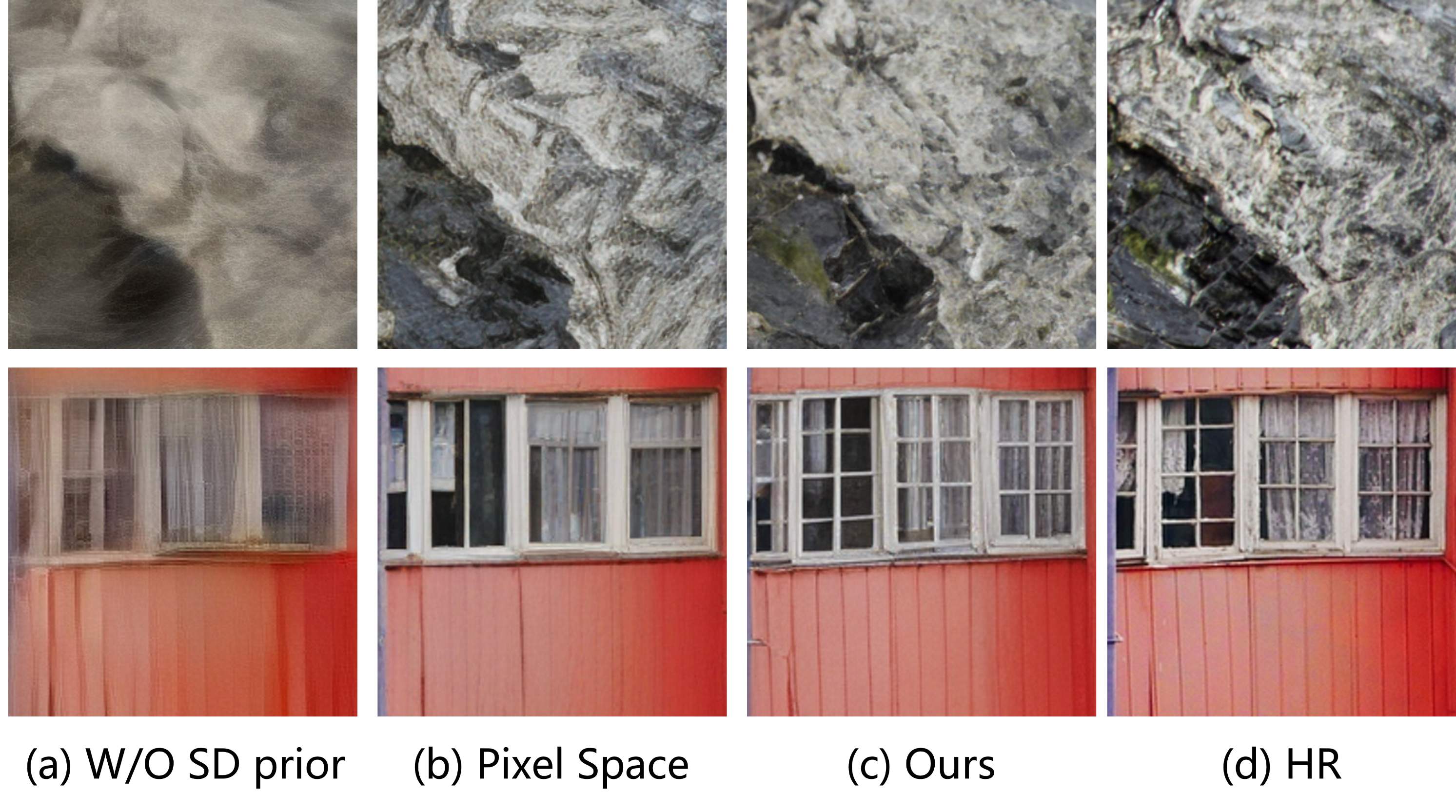}
    \caption{Visual comparisons of our method with rescaling in pixel space and without the SD prior.}
    \label{fig:vs_pixel}
\end{figure}

 As shown in Fig. \ref{fig:vs_pixel}, we conduct visual comparisons between rescaling in latent space and pixel space in two different scenarios. Our latent-space rescaling method is able to reconstruct more realistic textures and more accurate structural features. This indicates that latent-space rescaling operation can preserve sufficiently fine-grained contextual information about the HR image, such as structure, texture, and semantics.

\subsubsection{Necessity of SD Prior}
To validate that SD model provides rich priors about real-world HR images, we directly omit the perceptual enhancement part. As shown in Table \ref{tab:ablation_study_lsbir} (2nd row), the introduction of the SD priors can significantly improve performance of LPIPS and DISTS and greatly enhance the perceptual quality of the reconstruction results. Here, we compare the latent features before and after perceptual enhancement by decoding them to the pixel space using the VAE decoder, as shown in Fig. \ref{fig:vs_pixel}. It can be seen that the latent features before perceptual optimization exhibit noticeable color shifts and blurriness when mapped to the pixel domain. In contrast, the perceptually enhanced latent features display richer textures and more accurate color information. Therefore, leveraging the rich natural image priors in the pre-trained Stable Diffusion model can significantly enhance the visual quality of rescaled images.

\subsubsection{Effectiveness of DFRM Design}
\begin{figure}[tb]
\centering
\includegraphics[width=\linewidth]{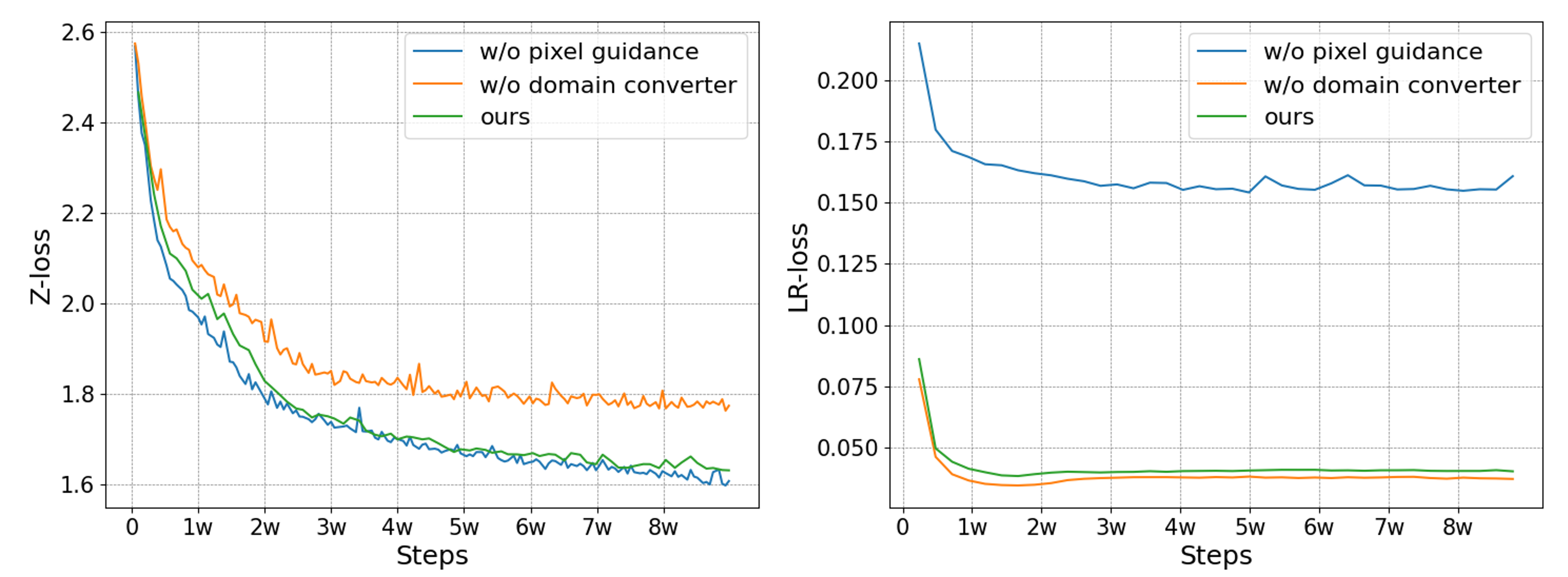}
\caption{The validation loss curves when removing a specific component from the DFRM.}
\label{fig:loss_curve}
\end{figure}

\begin{figure}[htb]
    \centering
    \includegraphics[width=\linewidth]{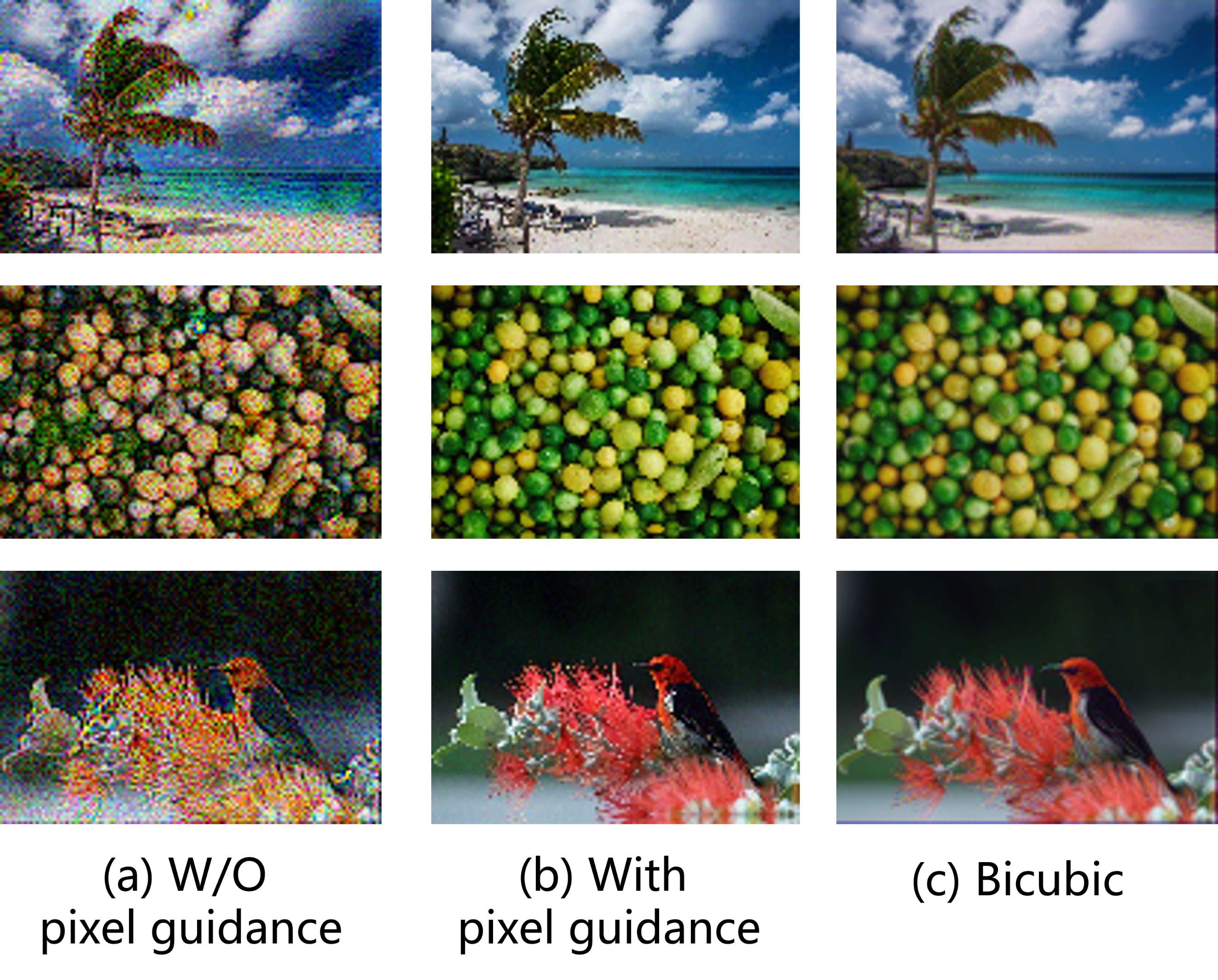}
    \caption{Visual comparisons of LR images with and without pixel guidance.}
    \label{fig:lr_quality_pg}
\end{figure}

Our DFRM includes two specific designs: the domain converter module and the pixel guidance module. To analyze their impact, we remove each component individually and plot the validation loss curves, as shown in Fig. \ref{fig:loss_curve}. It can be observed that the introduction of the INN can reduce reconstruction error. Although the pixel guidance module does not improve reconstruction performance, our experiments show that it can enhance the fidelity and visual quality of the LR images, as demonstrated by the lower guidance loss in Fig. \ref{fig:loss_curve}. As shown in Fig. \ref{fig:lr_quality_pg}, the LR images exhibit significant noise and color distortions in the absence of pixel guidance. However, the introduction of pixel guidance significantly alleviates these issues. Therefore, the design of our DFRM strikes a good balance between feature reconstruction and LR image quality.

\subsubsection{Effectiveness of Time Alignment}
\begin{figure}[tb]
\centering
\includegraphics[width=\columnwidth]{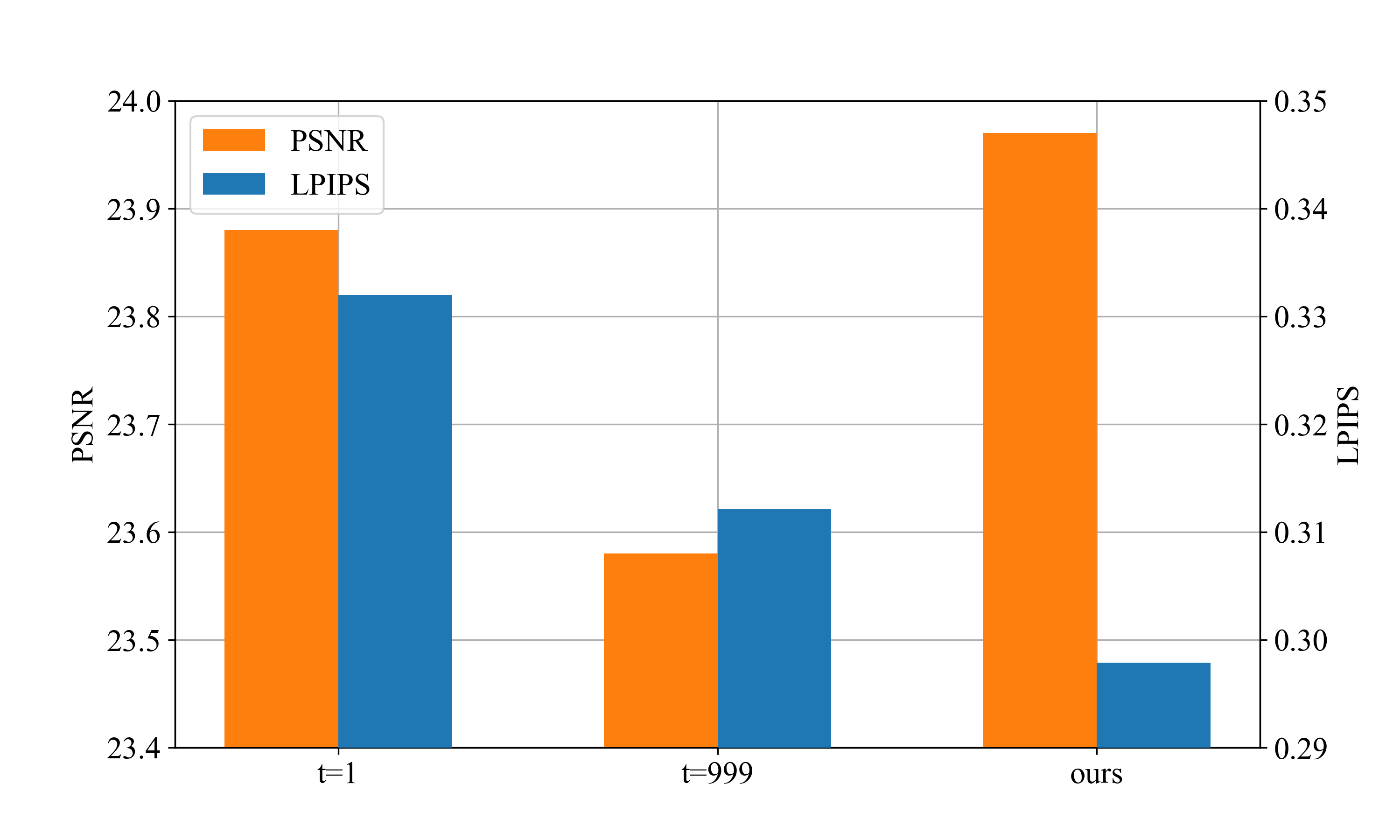}
\caption{Ablation study about time alignment.}
\label{fig:ablation_studt_ta}
\end{figure}

To validate the effectiveness of our proposed time-step alignment strategy, we retrain our model using fixed time-steps. This involves removing the TPM and time scheduler modules and employing a manually selected time-step to perform a single denoising step based on Eq. (\ref{equ:denoise}). Following the setting of previous work \cite{wu2024one}, we select fixed time-steps 1 and 999 for comparison. As shown in Fig. \ref{fig:ablation_studt_ta}, when the time-step is set to a smaller value, the model achieves higher fidelity but lower perceptual quality, and the situation is reversed when a larger time-step is chosen. In contrast, our time-step alignment mechanism can dynamically determine the level of generative capability based on the image content, thereby achieving optimal performance in both fidelity and perceptual quality. 

\begin{figure}[tb]
\centering
\includegraphics[width=\linewidth]{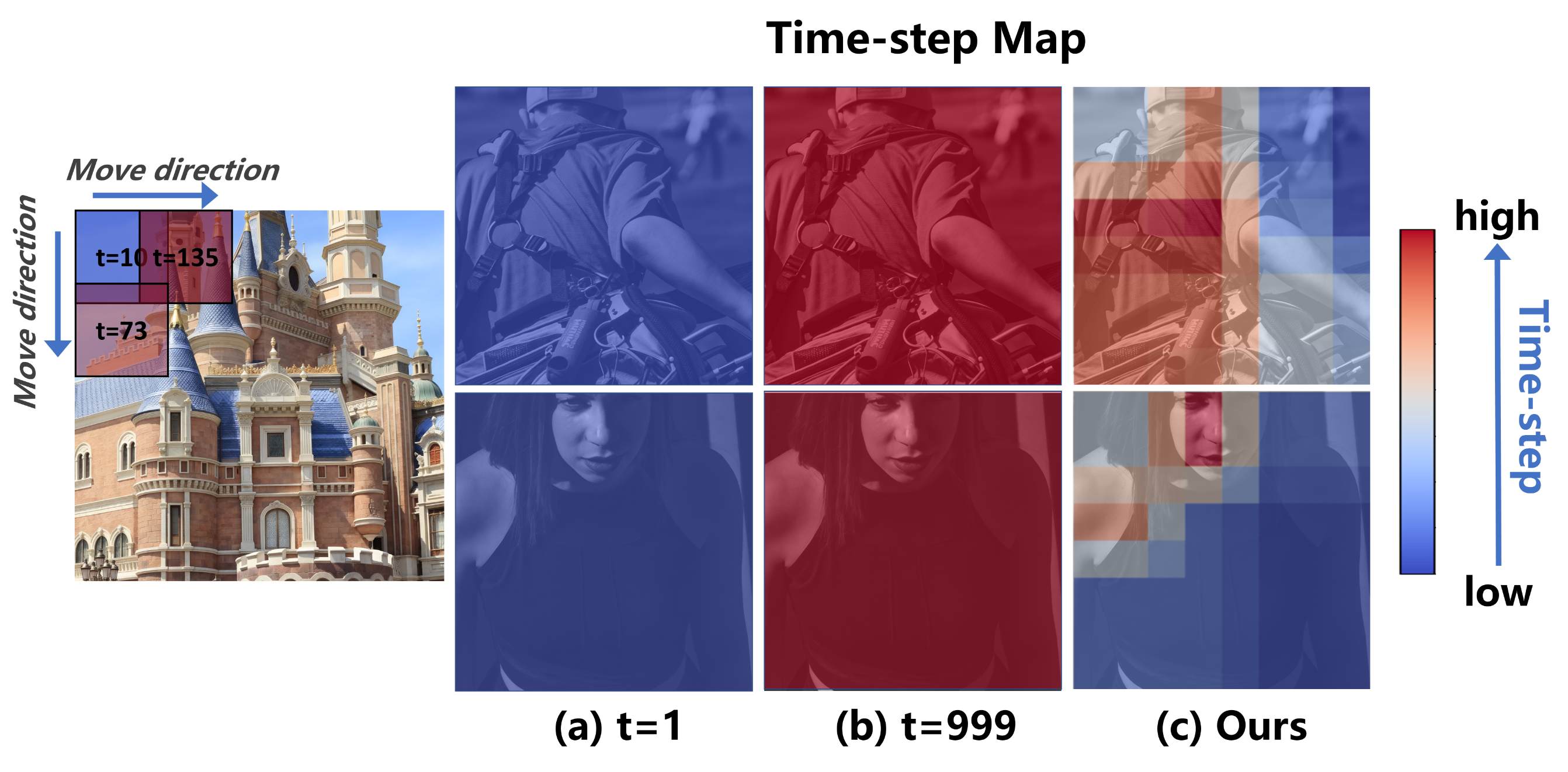}
\caption{Illustration of the tiled inference process and the predicted time-step map.}
\label{fig:time_step}
\end{figure}

In Fig. \ref{fig:time_step}, we present a schematic illustration of the tiled inference process along with the predicted time-step map. It can be observed that the time alignment mechanism effectively captures more complex regions in high-resolution images and assigns larger time steps to these areas.

\subsection{Discussions}
\subsubsection{Details about Tiled Inference}
When TADM is employed for rescaling ultra-high-resolution images, it often necessitates dividing the input image into multiple patches for separate processing. In the main paper, this is referred to as the tiled inference strategy. In Algorithm \ref{alg:tiled_inference}, we elaborate on the process of tiled inference. Unlike works such as StableSR \cite{wang2023exploiting}, our tiled inference algorithm can predict different time steps t for each image patch, thereby achieving dynamic allocation of generative capacity. 

\begin{algorithm}
\caption{Tiled Inference Process}
\label{alg:tiled_inference}
\renewcommand{\algorithmicrequire}{\textbf{Input:}}
\renewcommand{\algorithmicensure}{\textbf{Output:}}
\begin{algorithmic}[1]
\Require Input image $x$, latent encoder $\mathcal{E}$, decoupled feature rescaling module $\mathrm{DFRM}$, denoising U-Net $\epsilon_{\theta}$, time prediction module $\mathrm{TPM}$, time scheduler $\mathrm{TS}$, latent decoder $\mathcal{D}$, patch size $p$, stride length $s$
\Ensure Rescaled image $\hat{x}$, LR image $y$
\State $z=\mathcal{E}(x)$ \Comment{latent encoding}
\State $\hat{z}, y = \mathrm{DFRM}(z)$ \Comment{latent rescaling}
\State $[\hat{z}_{i}] = \mathrm{ToPatch}(\hat{z}, p, s)$ \Comment{split $\hat{z}$ to patches}
\State $z_{0}\_{list}=[]$ \Comment{initialize list of $z_{0}$}
\For {$\hat{z}_{i}$ in $[\hat{z}_{i}]$}
    \State $t_{i}=\mathrm{TPM}(\hat{z}_{i})$ \Comment{time-step prediction}
    \State $\epsilon_{i}=\epsilon_{\theta}(\mathrm{patch}, t_{i})$ \Comment{noise prediction}
    \State $z_{0}^{i}=\mathrm{TS}(\hat{z}_{i}, \epsilon, t_{i})$ \Comment{denoising by time scheduler}
    \State $z_{0}\_{list}.\mathrm{append}(z_{0}^{i})$
\EndFor
\State $z_{0}=\mathrm{Merge}(z_{0}\_{list})$ \Comment{merge patches of $z_{0}$}
\State $\hat{x}=\mathcal{D}(z_{0})$ \Comment{latent decoding}
\State
\Return $\hat{x}, y$ \Comment{output}
\end{algorithmic}
\end{algorithm}

The aforementioned algorithm encompasses two hyperparameters: the patch size and the stride length, both defined in the latent space. If the patch size is set too large, although inference efficiency may improve, the prediction of time steps would become overly sparse, leading to performance degradation. Conversely, if the patch size is set too small, the computational load will be significantly higher, and there might be a misalignment between the inference size and the pre-training image size of Stable Diffusion.

\begin{figure}[htb]
    \centering
    \includegraphics[width=\linewidth]{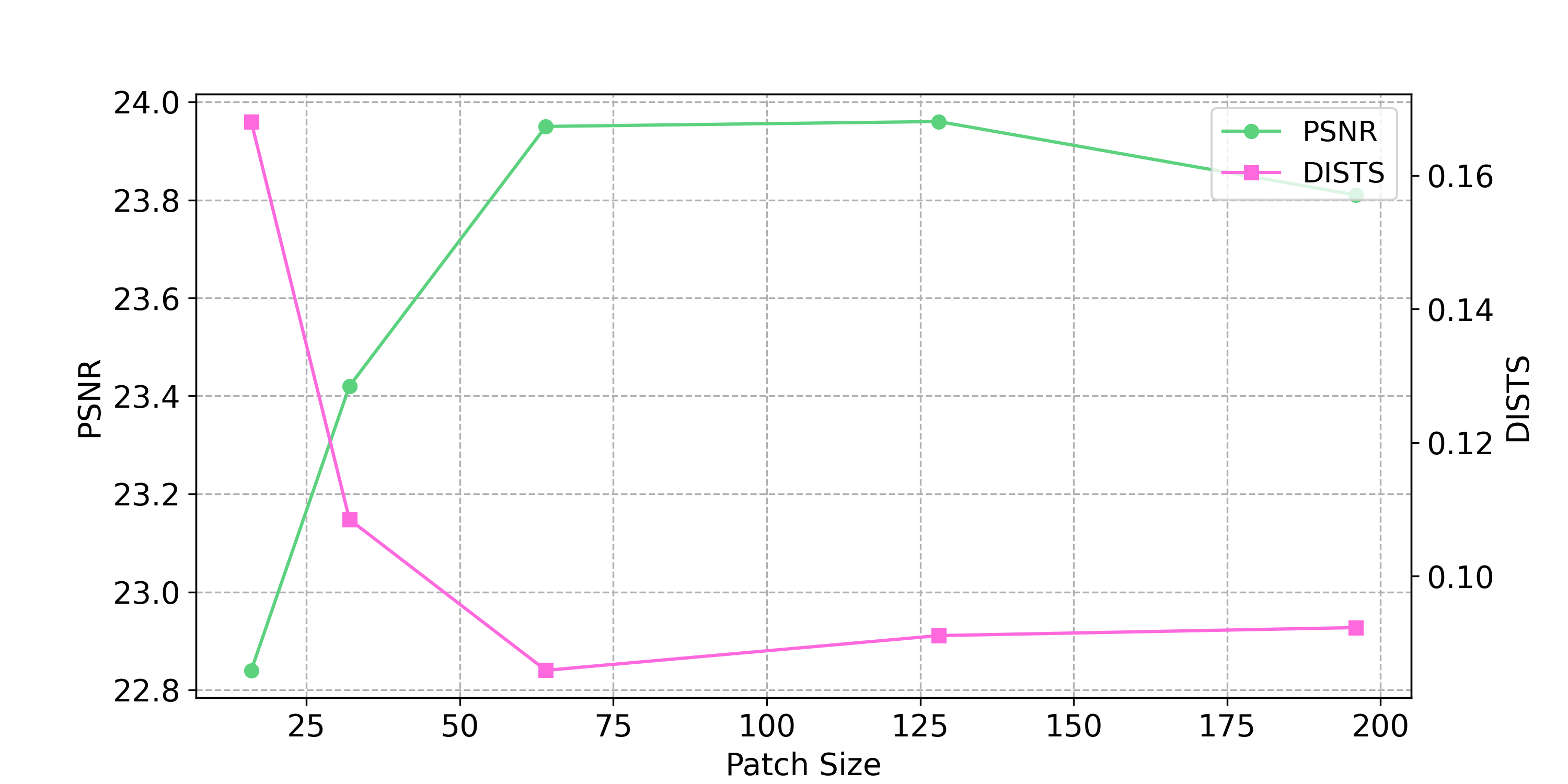}
    \caption{Ablation study about patch size.}
    \label{fig:patch_abla}
\end{figure}

Therefore, to achieve a trade-off between performance and inference efficiency, we conduct experiments on the patch size, as shown in Fig. \ref{fig:patch_abla}. It can be observed that the model achieves optimal performance when the patch size is between 60 and 120. Consequently, we select 96 as the patch size in our work. For the stride length, we adopt the value from previous works \cite{wang2023exploiting} and directly set it to 64.

\begin{figure}[htb]
    \centering
    \includegraphics[width=\linewidth]{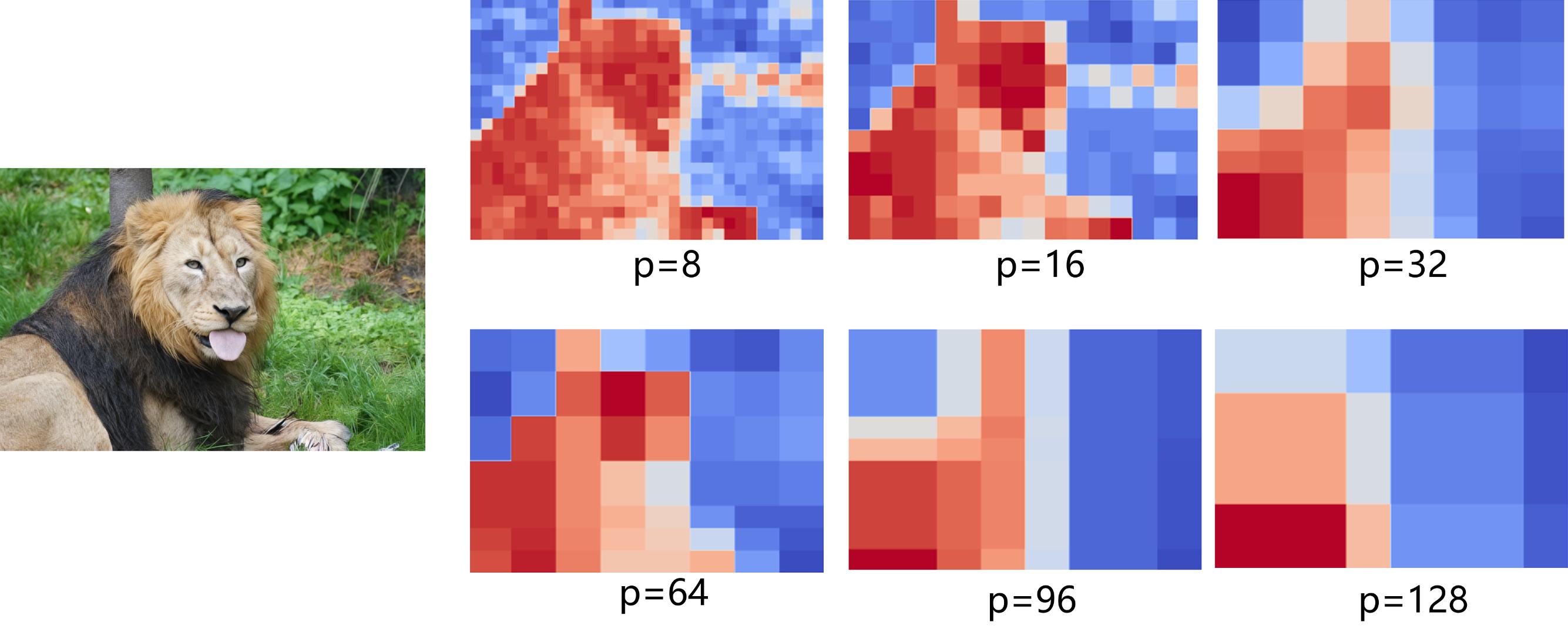}
    \caption{The time step predictions at different patch sizes.}
    \label{fig:timestep_dist}
\end{figure}

In Fig. \ref{fig:timestep_dist}, we present the time step predictions when employing different patch sizes during the tiled inference process. It can be observed that smaller patch sizes allow for finer-grained time step predictions. However, setting the size too small leads to misalignment with the pre-training image size of Stable Diffusion, resulting in increased computational load and performance degradation. In contrast, our selected patch size of 96 achieves a balance between performance and computational load, while also enabling relatively accurate prediction of the time step mask.

\subsubsection{Application to Image Compression}
\begin{figure}[htb]
    \centering
    \includegraphics[width=\linewidth]{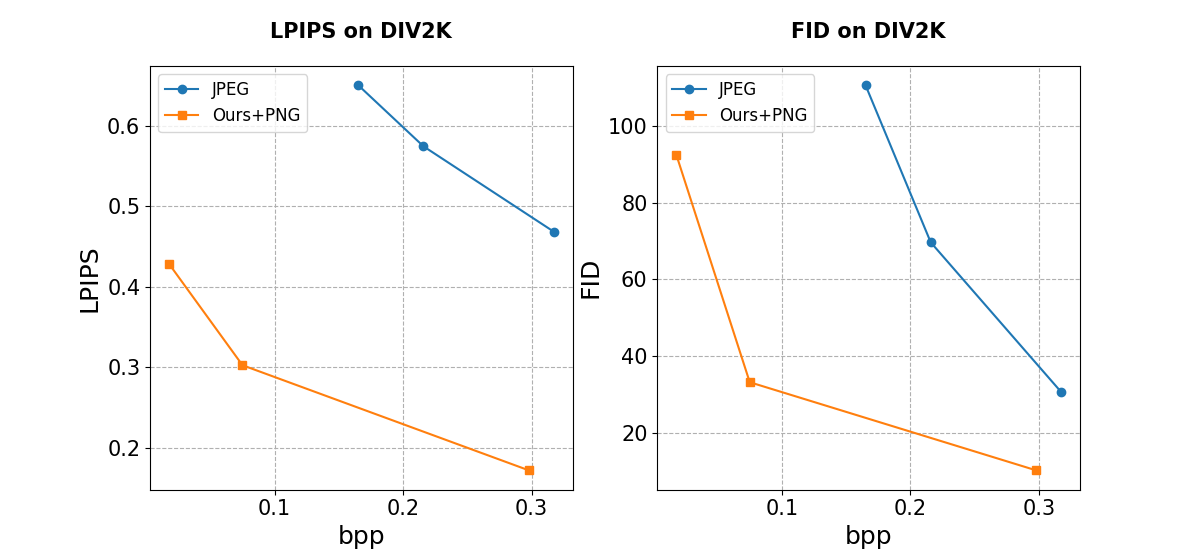}
    \caption{R-D curves of JPEG compression and our proposed model on the DIV2K validation set.}
    \label{fig:RD_curve}
\end{figure}

Our model can be combined with existing lossless image compression methods to achieve extreme compression ratios. To validate this, we use PNG to compress LR images and compare it with JPEG at various bit-rates. As shown in Fig. \ref{fig:RD_curve}, at the same bits per pixel (bpp), the image reconstruction quality of our model significantly surpasses that of the JPEG compression. These comparisons demonstrate the potential of our model for image compression.

\begin{figure}[htb]
    \centering
    \includegraphics[width=\linewidth]{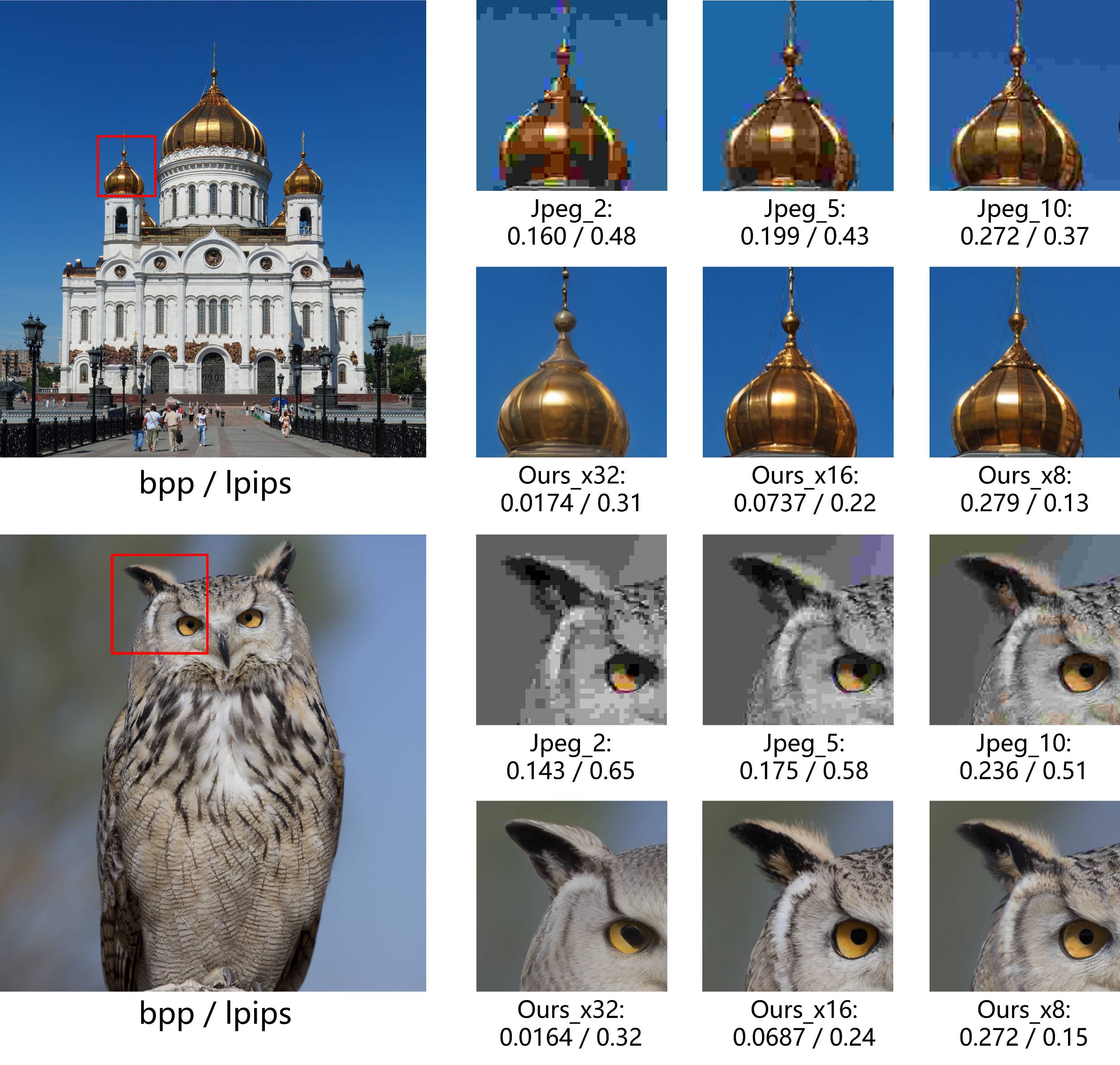}
    \caption{Visual  comparisons between our model and JPEG compression. bpp/LPIPS of the reconstructed images are shown below the images.}
    \label{fig:compare_jpeg}
\end{figure}

In Fig. \ref{fig:compare_jpeg}, we present the visual quality, the bpp and the LPIPS of the reconstructed images. Compared to JPEG compression, our method requires comparable or less storage space while achieving superior objective metrics. Additionally, JPEG compression tends to produce noticeable block artifacts and color distortions in the background regions of the images, whereas our model maintains superior reconstruction quality across all three scales.

\subsubsection{Inference Efficiency Comparison}
In Table \ref{tab:inference_cost}, we provide comparisons of inference time and memory consumption for representative methods. All tests are performed on an RTX 3090 using images with a resolution of $1024\times1024$. It can be observed that flow-based models (IRN and HCFlow) require less GPU memory consumption and enable fast inference. In contrast, StableSR, which is based on multi-step diffusion models, demands significantly more inference time and larger GPU memory consumption. Although VQIR requires less memory, the nearest neighbor search process in VQGAN results in longer inference time. For one-step diffusion-based method S3Diff, due to its degradation-guided LoRA which needs to dynamically predict LoRA parameters during inference, it requires longer ingerence time and higher memory consumption. In contrast, our proposed TADM demonstrates moderate efficiency and scalability compared to other methods. In the future, we can utilize distillation techniques to further reduce the model size.
\begin{table}[htb]
    \caption{Comparison of inference time and memory consumption.}
	\label{tab:inference_cost}
	\centering
    \resizebox{\linewidth}{!}{
    \setlength{\tabcolsep}{1.2mm}
	\begin{tabular}{ccccccc}
		\hline
         & IRN & HCFlow & VQIR & StableSR & S3Diff & Ours \\
        \hline
        Inference Time (s)& 0.359 & 0.363 & 0.897 & 135.2 & 2.186 & 0.711\\
        Memory Consumption (G)& 0.498 & 0.098 & 0.344 & 6.380 & 5.025 & 2.598 \\
        \hline
	\end{tabular}
        }
        \vspace{-0.4cm}
\end{table}

\subsubsection{Upper Bound of TADM}
Since our method performs the rescaling operation in the latent space of a pre-trained VAE, its performance upper boud is inherently constrained by the VAE. In Table \ref{tab:vae_rescale}, we present a performance comparison between VAE and two 4$\times$ rescaling models. For VAEs, we directly feed the high-resolution images to the encoder and decoder without performing any rescaling operations. This performance represents the upper bound of our method's performance across all rescaling factors. It can be observed that VAE significantly lags behind the latter. In the future, we could explore integrating stronger VAEs, such as those from SDXL, SD3, thereby enabling the TADM to be more applicable to small-scale rescaling tasks.
\begin{table}[htb]\footnotesize
    \caption{Comparison of VAE and 4$\times$ rescaling methods.}
	\label{tab:vae_rescale}
	\centering
	\begin{tabular}{cccc}
		\hline
         & IRN & HCFlow & VAE \\
        \hline
        PSNR & 35.07 & 35.23 & 29.01 \\
        SSIM & 0.9318 & 0.9346 & 0.8046 \\
        \hline
	\end{tabular}
\end{table}
\section{Conclusion}
We propose a novel framework called TADM to address issues such as insufficient texture and semantic inaccuracies in extreme image rescaling. Specifically, we design a DFRM to achieve bidirectional mapping between HR features and LR images. Then, we utilize a pre-trained SD model to perform a single denoising step on the rescaled features, enhancing its perceptual quality. Considering the non-uniform quality of the rescaled latent features, we propose a novel time-step alignment strategy to achieve a balance between fidelity and perceptual quality. Both qualitative and quantitative experiments demonstrate the superiority of our approach.
{
    \small
    \bibliographystyle{ieeenat_fullname}
    \bibliography{main}
}

\end{document}